\documentclass[runningheads]{llncs}

\usepackage{eccv}
\usepackage{eccvabbrv}
\usepackage{graphicx}
\usepackage{booktabs}
\usepackage[accsupp]{axessibility}
\usepackage{amsmath,amssymb,amsfonts}
\usepackage{multirow}
\usepackage{caption}
\usepackage{subcaption}
\usepackage{xcolor}
\usepackage{algorithm}
\usepackage{algorithmic}
\usepackage{bm}
\usepackage{enumitem}
\usepackage{stfloats}
\usepackage{hyperref}
\usepackage{orcidlink}

\newcommand{\vx}{\mathbf{x}}
\newcommand{\vz}{\mathbf{z}}
\newcommand{\vm}{\mathbf{m}}
\newcommand{\vf}{\mathbf{f}}
\newcommand{\R}{\mathbb{R}}
\newcommand{\Rbev}{R_{\mathrm{BEV}}}
\newcommand{\Rpose}{R_{\mathrm{pose}}}

\begin{document}

\title{ModTrack: Sensor-Agnostic Multi-View Tracking via Identity-Informed PHD Filtering with Covariance Propagation}

\titlerunning{ModTrack}


\author{
Aditya Iyer$^{*\dagger}$ \and
Jack Roberts$^{*\dagger}$ \and
Nora Ayanian$^{\dagger\ddagger}$
}

\authorrunning{A.~Iyer et al.}

\institute{$^\dagger$Department of Computer Science \ \ $^\ddagger$School of Engineering \\ Brown University, Providence, RI, USA\\
\email{\{aditya\_iyer, jack\_roberts, nora\_ayanian\}@brown.edu}\\
$^{*}$Equal contribution.}
\maketitle

\begin{abstract}
Multi-View Multi-Object Tracking (MV-MOT) aims to localize and maintain consistent identities of objects observed by multiple sensors. This task is challenging, as viewpoint changes and occlusion disrupt identity consistency across views and time. Recent end-to-end approaches address this by jointly learning 2D Bird's Eye View (BEV) representations and identity associations, achieving high tracking accuracy. However, these methods offer no principled uncertainty accounting and remain tightly coupled to their training configuration, limiting generalization across sensor layouts, modalities, or datasets without retraining. We propose ModTrack, a modular MV-MOT system that matches end-to-end performance while providing cross-modal, sensor-agnostic generalization and traceable uncertainty. ModTrack confines learning methods to just the \textit{Detection and Feature Extraction} stage of the MV-MOT pipeline, performing all fusion, association, and tracking with closed-form analytical methods. Our design reduces each sensor's output to calibrated position-covariance pairs $(\vz, R)$; cross-view clustering and precision-weighted fusion then yield unified estimates $(\hat{\vz}, \hat{R})$ for identity assignment and temporal tracking. A feedback-coupled, identity-informed Gaussian Mixture Probability Hypothesis Density (GM-PHD) filter with HMM motion modes uses these fused estimates to maintain identities under missed detections and heavy occlusion. ModTrack achieves 95.5 IDF1 and 91.4 MOTA on \textit{WildTrack}, surpassing all prior modular methods by over 21 points and rivaling the state-of-the-art end-to-end methods while providing deployment flexibility they cannot. Specifically, the same tracker core transfers unchanged to \textit{MultiviewX} and \textit{RadarScenes}, with only perception-module replacement required to extend to new domains and sensor modalities.
\end{abstract}

\begin{figure*}[!t]
\centering
\includegraphics[width=1\textwidth]{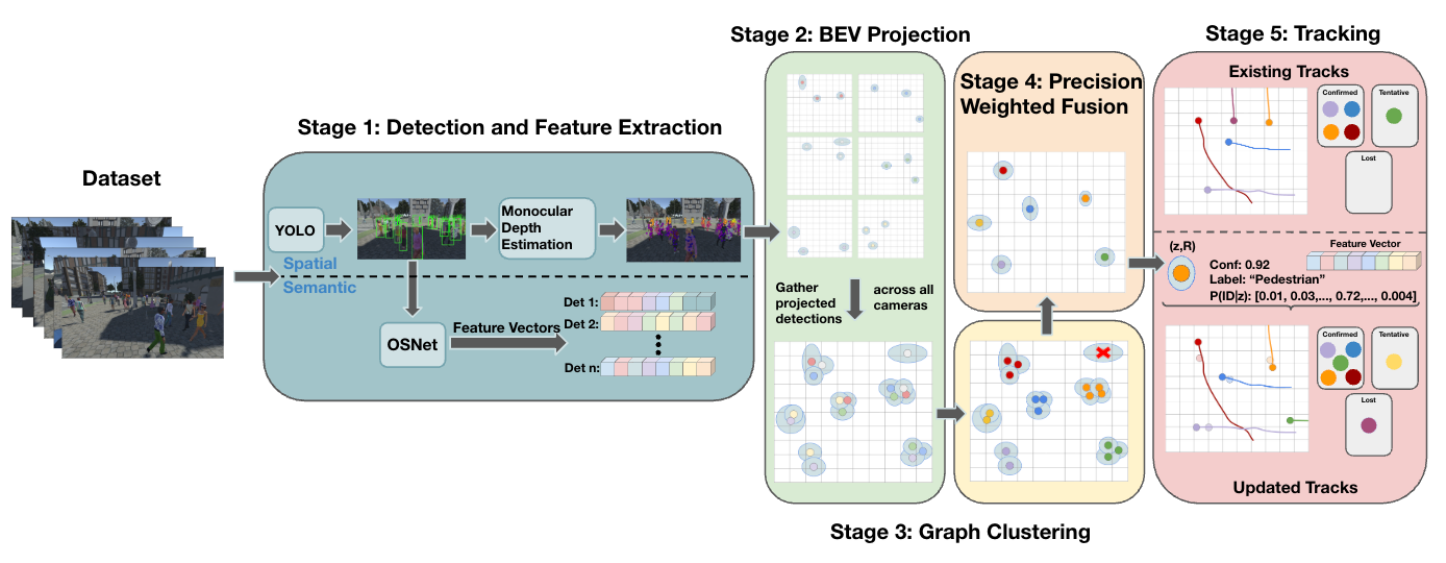}
\caption{\textbf{ModTrack pipeline.} Per-camera detections and appearance features (Stage~1) are projected to BEV with Jacobian-propagated covariance (Stage~2), associated via $\chi^2$ graph clustering (Stage~3), and precision-weighted into fused $(\hat{\vz},\hat{R})$ pairs (Stage~4). An identity-informed GM-PHD filter (Stage~5) maintains persistent identities, with track predictions feeding back to guide identity assignment (dashed). Neural modules (blue) are confined to Stage~1; all downstream stages are purely analytical.}
\label{fig:pipeline}
\end{figure*}

\section{Introduction}
\label{sec:intro}
Multi-view multi-object tracking (MV-MOT) underpins applications from surveillance to collaborative robotics. This task is challenging because objects observed from different viewpoints vary significantly in appearance, scale, and visibility, while occlusion and missed detections disrupt identity consistency across both views and time.

Recent end-to-end approaches~\cite{earlybird, mcblt, tracktacular, scfusion}
achieve high accuracy by jointly learning Bird's Eye View (BEV) representations and data
association, but they provide no interpretable uncertainty estimates and
remain tightly coupled to the sensor configuration seen during training,
making adaptation to new sensors or modalities difficult without retraining or architectural modification.

Classical random finite set (RFS) filters provide principled multi-target
estimation with closed-form covariance propagation.
However, the standard PHD filter does not maintain persistent identities~\cite{phd}.
Labeled extensions~\cite{vovo2014} recover identities but require
combinatorial hypothesis enumeration, and existing multi-view RFS trackers do not propagate measurement covariances from perception through cross-view fusion and tracking, limiting uncertainty interpretability and downstream calibration~\cite{ong}.

We propose ModTrack, a modular MV-MOT system that rivals end-to-end accuracy while providing interpretable, traceable uncertainty, by confining learning to just the \textit{Detection and Feature Extraction} stage of the pipeline (\Cref{fig:pipeline}) and performing all fusion, association, and tracking with closed-form analytical methods. The tracking backend uses the GM-PHD predict-update recursion as an analytical backbone for state estimation and component management, augmented with deterministic identity assignment and lifecycle logic for persistent track continuity~\cite{voba2006}. Supplementary Material~X validates that these modifications improve identity consistency while preserving state-estimation accuracy. ModTrack adapts to new sensor configurations at inference time seamlessly, avoiding the retraining or engineering accomodations typically required by end-to-end approaches. It also extends prior RFS trackers by operating directly in calibrated BEV space with metric covariance propagation, narrowing the accuracy gap to learned methods in high-clutter scenes. Across domains, all sensor measurements are represented as calibrated position-covariance pairs $(\vz,R)$, so the same analytical core runs unchanged on WildTrack~\cite{wildtrack} (7 real cameras), MultiviewX~\cite{multiviewx} (6 synthetic cameras), and RadarScenes~\cite{radarscenes} (4 automotive radars), requiring only perception-module replacement. On WildTrack, ModTrack achieves 95.5 IDF1 and 91.4 MOTA, surpassing all prior modular methods by over 21 points and rivaling state-of-the-art end-to-end methods.

The contributions of this work are threefold.
\begin{enumerate}[leftmargin=*]

\item \textbf{Identity-informed GM-PHD filter with closed-loop propagation.}
We augment the GM-PHD filter with persistent identity labels, HMM motion modes, and appearance features. Track predictions feed back as association priors into each PHD update, yielding a +8.9 IDF1 gain over spatial information-only association.

\item \textbf{Analytical covariance chain.}
A fully interpretable uncertainty pipeline propagates monocular depth variance through Jacobian BEV projection, $\chi^2$-calibrated graph clustering, and common-mode-aware precision-weighted fusion, with every covariance traceable to physical sensor characteristics.

\item \textbf{Sensor-agnostic deployment flexibility.}
The identical tracker core operates unchanged on WildTrack, MultiviewX, and RadarScenes with only perception-module replacement, and we validate robustness to sensor dropout over all $\binom{7}{k}$ camera subsets of WildTrack.

\end{enumerate}

\section{Related Work}
\label{sec:related}

\textbf{End-to-End and Learned Multi-View Tracking.}
Early multi-camera systems used single-view trackers~\cite{chen2018multi} independently per camera and reconciled identities across views using appearance cues and epipolar constraints. Learned \newline
re-identification embeddings improved cross-view matching~\cite{zheng2015scalable, wei2018person}.
More recent approaches perform joint reasoning directly in BEV. EarlyBird~\cite{earlybird} projects per-camera features to a shared
ground plane before joint detection and tracking, while later work introduces
hierarchical graph reasoning (MCBLT~\cite{mcblt}), sparse cross-view
transformers (SCFusion~\cite{scfusion}), view-aware attention
mechanisms~\cite{alturki2025attention}, and trajectory-level motion–appearance
aggregation (MVTrajecter~\cite{mvtrajecter}, TrackTacular~\cite{tracktacular}).
Related formulations treat multi-view association as a learned graph problem
(UMPN~\cite{onegraph}) or support variable camera configurations for BEV
detection (MIC-BEV~\cite{micbev}). Despite strong performance, these methods
typically embed sensor geometry and fusion within learned parameters and do
not expose explicit uncertainty estimates.
\\
\textbf{Random Finite Set Tracking.}
Random finite set (RFS) formulations provide a principled Bayesian framework for multi-target estimation under clutter and missed detections. The GM-PHD filter~\cite{voba2006} offers closed-form multi-target state estimation with linear complexity in the number of mixture components, but does not maintain persistent identity labels. The GLMB filter~\cite{vovo2014} restores labeled tracking with full Bayesian data association, yet requires enumerating association hypotheses whose number grows combinatorially. The LMB filter~\cite{reuter2014mb} offers a tractable approximation but, like GLMB, carries only statistical labels without appearance integration or feedback from track state to measurement assignment. MV-GLMB-OC~\cite{ong} extends GLMB to MV-MOT in pedestrian settings, incorporating occlusion reasoning across cameras, but operating in image-space bounding boxes and is unable to propagate metric covariances in a unified world frame. While RFS-based methods provide rigorous uncertainty handling, they have not been combined with BEV-based geometric fusion or scalable identity reasoning as used in contemporary learned MV-MOT systems.

ModTrack bridges learned BEV-based tracking and classical RFS filtering. By combining precision-weighted multi-view fusion with an RFS-style identity-informed GM-PHD backend, it preserves explicit probabilistic state estimates while remaining competitive with modern learned MV-MOT systems.

\section{Methodology}
\label{sec:method}

\subsection{Detection and Feature Extraction}
\label{sec:detection}

ModTrack's detection and feature extraction stage is sensor-agnostic: any front-end that produces calibrated BEV position-covariance pairs $(\vz, R)$ can drive the identical downstream tracking pipeline. We describe two perception modules in this work, a monocular vision front-end and an analytical radar front-end, to validate this claim across modalities. Empirical validation on additional modalities remains future work.

\subsubsection{Camera Front-End (WildTrack, MultiviewX).}
At each time step, \newline YOLOv11x~\cite{yolo11_ultralytics} predicts bounding boxes, class labels, and confidence scores for each camera.
For each detection, we estimate metric depth by precision-weighting two complementary geometric cues: a ground-plane ''footpoint'' back-projection and a bounding-box height prior under the pinhole camera model. We clarify ''footpoint'' back-projection does not require explicit footpoint localization but rather approximates the footpoint as the lowest central point of the bounding box. A learned monocular depth network~\cite{philion2020lift} serves as an outlier check, inflating the fused variance when it confidently disagrees with the geometric estimate. The resulting depth $\hat{d}$ and variance $\sigma_d^2$ are passed to the BEV projection stage (\Cref{sec:measurement}); full derivations and parameter choices are provided in Supplementary Material sections A and F, respectively. 

OSNet-AIN~\cite{zhou2021osnetain} then processes each bounding box and computes feature vectors for each detection. These serve as appearance/semantic identity cues used to augment the PHD filter.

\subsubsection{Radar Front-End (RadarScenes).}
\label{sec:radar_frontend}
Each radar measurement provides range $r$ and azimuth $\theta$ with known uncertainties $(\sigma_r, \sigma_\theta)$, where $\theta$ is measured from the forward (longitudinal) sensor axis so that $x_s = r\sin\theta$ and $y_s = r\cos\theta$ in the sensor-local frame.
The Jacobian of this polar-to-Cartesian transform,
\begin{equation}
J_\text{radar} = \begin{bmatrix} \sin\theta & r\cos\theta \\ \cos\theta & -r\sin\theta \end{bmatrix},
\label{eq:radar_jacobian}
\end{equation}
yields sensor-local Cartesian covariance $R_s = J_\text{radar}\,\text{diag}(\sigma_r^2, \sigma_\theta^2)\,J_\text{radar}^\top$. A rotation $R_\psi$ (ego-vehicle yaw plus sensor mount angle) transforms to the global BEV frame:
\begin{equation}
\Rbev = R_\psi\,R_s\,R_\psi^\top + \Rpose,
\label{eq:radar_rbev}
\end{equation}
Object class and confidence scores are derived from Doppler velocity and radar cross-section via ad-hoc thresholds, replacing YOLO with zero learned parameters.
DBSCAN pre-clustering~\cite{ester1996density} groups per-point detections into object-level measurements. This produces the same $(\vz, R)$ interface without any semantic information.

\subsection{Ground-Plane Projection with Jacobian Uncertainty Propagation}
\label{sec:measurement}
This subsection applies to the camera front-end (\Cref{sec:detection}); the radar front-end (\Cref{sec:radar_frontend}) produces $(\vz, R)$ pairs directly and bypasses ground-plane projection.

Given a detection at pixel $(u, v)$ in camera $c$ with estimated depth $\hat{d}$ and variance $\sigma_d^2$, we back-project
$\mathbf{r}_c = K^{-1}[u, v, 1]^\top$, map the ray in the camera frame into the world frame at the depth $\hat{d}$, and orthogonally project it onto the ground plane:
\begin{equation}
\mathbf{X}_w = R_c\,(\hat{d}\,\mathbf{r}_c) + \mathbf{t}_c, \quad
\mathbf{X}_{\mathrm{proj}} = \mathbf{X}_w + \left(d_{\mathrm{plane}} - \mathbf{n}^\top \mathbf{X}_w\right)\mathbf{n},
\end{equation}
and define the BEV position $\vz_{\mathrm{BEV}} = \mathbf{X}_{\mathrm{proj},\,1:2}$.

We then propagate the scalar depth uncertainty through the projection using the Jacobian with respect to $\hat{d}$, producing a rank-1 BEV covariance term $R_{\mathrm{indep}} = \sigma_d^2\,J\,J^\top$. The full derivation, including the closed-form Jacobian and the role of the horizontal-plane assumption, is given in Supplementary Material (Supp. Material) A.1.
We model the BEV measurement covariance as the sum of the depth-propagated term and an isotropic calibration/ego-pose term, $\Rbev = R_{\mathrm{indep}} + \Rpose$, where $\Rpose = \sigma_{\mathrm{pose}}^2 I_2$ and $\sigma_{\mathrm{pose}}$ is set per dataset (Supp. Material Table 6); Supp. Material A.1 gives the eigenvalue flooring used to avoid overconfident gating.
This defines the \emph{sensor-agnostic $(\vz, R)$ interface} for the camera front-end.

\subsection{$\chi^2$ Graph Clustering}
\label{sec:clustering}
In order to group $(\vz, R)$ pairs from different sensors associated with the same object we construct an association graph where edge weights encode the $\chi^2$ consistency (survival) probability that two detections originate from the same target.
For detections $p_i$ and $p_j$ from distinct sensors, the Mahalanobis distance under summed covariances is
\begin{equation}
d^2_{ij}(p_i, p_j) = (\vz_i - \vz_j)^\top (R_i + R_j)^{-1} (\vz_i - \vz_j).
\label{eq:mahal}
\end{equation}
Under the null hypothesis that both observations originate from the same physical location, $d^2_{ij}$ follows a $\chi^2$ distribution with 2 degrees of freedom.
The resulting $\chi^2$ consistency score is $P_{\mathrm{same}}(p_i, p_j) = 1 - F_{\chi^2_2}(d^2_{ij})$, where $F_{\chi^2_2}$ is the $\chi^2$ CDF.
Edges with $P_{\mathrm{same}} > \tau_P$ (corresponding to the $\chi^2_{2,0.99}{=}9.21$ gate at $\tau_P{=}0.01$) are retained; a hard Euclidean cutoff $\tau_{\mathrm{euc}}$ prevents transitive chaining. Following ByteTrack-style~\cite{zhang2022bytetrack} cascaded confidence processing, we first cluster high-confidence esimations, then cluster low-confidence estimations in a second pass only for tracks left unmatched after the first pass.
Because transitive closure can place multiple estimates from the same sensor into one component, we additionally split each estimate into sensor-unique groups (at most one estimate per sensor, selecting higher-confidence estimates first) before downstream fusion. This is based on the assumption that the object detector does not produce duplicate detections for a single object.
We retain only connected components supported by at least two distinct sensors; singleton components are discarded before fusion. This is a pedestrian dataset specific design choice. We observed that the vast majority of single-view clusters were false positives due to the highly overlapping camera coverage making single-view detections rare. For RadarScenes all DBSCAN clusters, including singletons, reach the PHD tracker.

\subsection{Precision-Weighted Multi-View Fusion}
\label{sec:fusion}

For each cluster with $M$ sensor measurements $\{(\vz_m, R_m)\}_{m=1}^{M}$, we decompose each covariance into an independent term and a shared common-mode term (calibration/ego-pose), $R_m = R_{\mathrm{indep},m} + \Rpose$, and fuse only the independent components (when the shared term is available and consistent across measurements; otherwise $\Rpose{=}0$):
\begin{align}
P_{\mathrm{indep}}^{-1} &= \textstyle\sum_{m=1}^{M} R_{\mathrm{indep},m}^{-1}, \label{eq:pfused}\\
\hat{\vz}_{\mathrm{fused}} &= P_{\mathrm{indep}} \textstyle\sum_{m=1}^{M} R_{\mathrm{indep},m}^{-1}\,\vz_m, \label{eq:xfused}\\
P_{\mathrm{fused}} &= P_{\mathrm{indep}} + \Rpose. \label{eq:pfused_common}
\end{align}
The fused estimate $(\hat{\vz}, \hat{R}) \triangleq (\hat{\vz}_{\mathrm{fused}},\, P_{\mathrm{fused}})$ remains order-independent; importantly, only the independent perceptual uncertainty shrinks with additional sensors, while the shared calibration/ego-pose term is added back once and therefore does not spuriously vanish.
A confidence-weighted appearance feature is pooled across cluster members when semantic information is available:
\begin{equation}
\vf_{\mathcal{C}} = \frac{\sum_{m=1}^{M} \beta_m \cdot \hat{\vf}_m}{\left\|\sum_{m=1}^{M} \beta_m \cdot \hat{\vf}_m\right\| + \epsilon},
\label{eq:feature_pooling}
\end{equation}
where $\hat{\vf}_m = \vf_m / \|\vf_m\|$ is the $L_2$-normalized appearance feature and $\beta_m$ is the detection confidence.

\subsection{Identity-Informed GM-PHD Filter and Closed-Loop Identity Assignment}
\label{sec:phd}

\subsubsection{GM-PHD State Definition and Tracking Loop.}
We build on the GM-PHD predict-update recursion~\cite{voba2006} as the backbone for state estimation and component management, augmenting it with persistent identity labels, appearance features, HMM motion modes, and lifecycle heuristics for practical identity maintenance. These pragmatic extensions prioritize robust identity tracking over strict adherence to the RFS-theoretic PHD recursion. The $j$-th Gaussian component at time step $k$ is:
\begin{equation}
\mathcal{C}_k^{(j)} = \bigl\{w_k^{(j)},\; \vm_k^{(j)},\; P_k^{(j)},\; \mathrm{ID}_k^{(j)},\; s_k^{(j)},\; \vf_k^{(j)}\bigr\},
\label{eq:component}
\end{equation}
where $w$ is the PHD weight, $\vm{=}[x,y,\dot{x},\dot{y}]^\top$ is the BEV state, $P{\in}\R^{4\times4}$ is the state covariance, $\mathrm{ID}$ is the persistent identity, $s{\in}\{1,2,3\}$ is the HMM motion mode, and $\vf$ is the pooled appearance feature.

We distinguish between \emph{components} and \emph{tracks}: a component $j$ is a single Gaussian hypothesis (typically corresponding to one motion-mode hypothesis), while a track $t$ denotes a persistent identity label and may be represented by multiple components $\mathcal{J}_k(t)\triangleq\{j:\mathrm{ID}_k^{(j)}=t\}$ within a frame.

At each frame, the method follows the same loop: (i) export one representative GM-PHD component state per track to the fusion stage, (ii) assign each fused cluster both a hard identity and a soft identity distribution using that exported pool, (iii) predict the existing GM-PHD components forward by one time step and append identity-informed birth components, (iv) perform track-level association and Kalman update, and then (v) prune, merge, and apply lifecycle transitions before exposing the next frame's track pool.

\subsubsection{Closed-Loop Identity Prior.}
\label{sec:identity}
The fusion stage and the PHD filter communicate in a \emph{closed loop}. Before processing new measurements, the filter exports one representative component per visible track,
$\mathcal{T}_k = \{(t, \vm_k^{(j_t^\star)}, P_k^{(j_t^\star)}, s_k^{(j_t^\star)}, \vf_k^{(j_t^\star)})\}_{t=1}^{T_k}$,
where $j_t^\star = \arg\max_{j \in \mathcal{J}_k(t)} w_k^{(j)}$.
For each fused estimate $(\hat{\vz}_i, \hat{R}_i)$ and representative component $t$, a gated Mahalanobis cost drives Hungarian matching~\cite{kuhn1955hungarian}:
\begin{equation}
\text{Cost}_{i,t} =
\begin{cases}
d_{i,t}^2, & d_{i,t}^2 < \tau_{\text{gate}} \\
+\infty, & \text{otherwise}
\end{cases} \, , \text{Cost}_{\text{new}} = \tau_{\text{new}}
\label{eq:cost}
\end{equation}
Here $d_{i,t}^2$ is computed as in \Cref{eq:mahal} with $p_i \triangleq (\hat{\vz}_i,\hat{R}_i)$ and $p_t \triangleq (\vz_t, R_t)$, where $(\vz_t, R_t)$ is the predicted BEV position and covariance from the $\vm, P$ entries of the representative component for track $t$.

A heuristic spatial identity score provides a soft preference toward nearby tracks. Rather than a full Bayesian posterior, we apply the same $\chi^2_2$ gate as \Cref{eq:cost}: the soft distribution is formed only over tracks with $d_{i,t}^2 < \tau_{\text{gate}}$ and then renormalized:
\begin{equation}
g_{i,t} =
\exp\!\left(-\frac{d_{i,t}^{2}}{2\,\sigma_{\mathrm{spatial}}^{2}}\right),
\quad
\pi^{\mathrm{geo}}_{i,t} =
\frac{\exp(g_{i,t} / \tau_{\mathrm{geo}})}
{\sum_{t'} \exp(g_{i,t'} / \tau_{\mathrm{geo}})} .
\label{eq:identity_prob}
\end{equation}
where $\sigma_{\mathrm{spatial}} = 1.0$ is a fixed spatial bandwidth and $\tau_{\mathrm{geo}}$ controls the sharpness of the softmax.
$\pi^{\mathrm{geo}}_{i,t}$ denotes a normalized spatial proximity score for assigning measurement i to track t.
The hard Hungarian identity initializes the candidate persistent label for the fused cluster, while the soft distribution re-enters the tracker through the association cost in \Cref{eq:hungarian}. The Hungarian-assigned identity receives nonzero mass before renormalization, to ensure each measurement provides a valid distribution.

\subsubsection{Prediction and Identity-Informed Birth.}
\label{sec:hmm}
Three motion modes share the linear state-transition structure $\vx_{k+1} = F_s\,\vx_k + \mathbf{w}_k$ with $\mathbf{w}_k \sim \mathcal{N}(0, Q_s)$: \emph{stationary} ($s{=}1$) damps velocity, \emph{constant velocity} ($s{=}2$) uses the standard CV transition, and \emph{maneuvering} ($s{=}3$) uses the same transition with elevated process noise $Q_3 \gg Q_2$~\cite{blom1988imm}. Mode transitions follow a class-specific Markov chain $\Pi^{(c)}_{ij} = P(s_{k+1}{=}j \mid s_k{=}i)$.

Prediction and birth are executed as a single stage. We first propagate the surviving components through the HMM motion model,
\begin{align}
w_{k|k-1}^{(j,s')} &= p_S^{(c)} \cdot \Pi^{(c)}_{s_k^{(j)},s'} \cdot w_k^{(j)}, \label{eq:wpred}\\
\vm_{k|k-1}^{(j,s')} &= F_{s'}\,\vm_k^{(j)}, \label{eq:mpred}\\
P_{k|k-1}^{(j,s')} &= F_{s'}\,P_k^{(j)}\,F_{s'}^\top + Q_{s'}^{(c)}. \label{eq:ppred}
\end{align}
and then append identity-aware birth components from the current fused clusters whose hard identity assignment does not match an existing predicted track and whose confidence exceeds the birth threshold $\tau_{\mathrm{birth}}$,
\begin{equation}
\vm_{\mathrm{birth}}^{(s)} = \begin{bmatrix} \hat{\vz}_{\mathcal{C}} \\ \mathbf{0} \end{bmatrix},\;
P_{\mathrm{birth}}^{(s)} = \begin{bmatrix} \hat{R}_{\mathcal{C}} & 0 \\ 0 & (\sigma_v^{(c)})^{2} I_2 \end{bmatrix},\;
w_{\mathrm{birth}}^{(s)} = \beta_{\max} \cdot \bar{\Pi}_{1,s},
\label{eq:birth}
\end{equation}
where $\sigma_v^{(c)}$ is the class-specific birth velocity prior, $\bar{\Pi}_{1,:}$ is the stationary-mode prior, and $\beta_{\max}$ is the maximum detection confidence in the cluster. Multiple mode hypotheses are retained through association and update, then collapsed during post-update component management.

\subsubsection{Track-Level Association and Update.}
\label{sec:hungarian}
At this point, the predicted pool contains both propagated hypotheses from existing tracks and newly initialized tentative birth hypotheses. These predicted components are associated against the current fused measurements from the same frame. Association is performed at the \emph{track level} rather than the component level so that multiple HMM modes of the same identity do not compete for different measurements within the same frame.

\textbf{Cost construction.}
For each track component $j$ and measurement $i$, the base cost is the Gaussian negative log-likelihood:
\begin{equation}
c_{ij}^{\mathrm{base}} = \tfrac{1}{2}\bigl(d^2_{ij} + \log|S_k^{(j,i)}|\bigr) - \log p_D^{(c)},
\label{eq:nll_cost}
\end{equation}
where $d^2_{ij} = (\vz_k^{(i)} - H\vm_{k|k-1}^{(j)})^\top (S_k^{(j,i)})^{-1} (\vz_k^{(i)} - H\vm_{k|k-1}^{(j)})$ and \newline
$S_k^{(j,i)} = HP_{k|k-1}^{(j)}H^\top + R_k^{(i)}$ is the innovation covariance ($H{=}[I_2 \mid 0]$). We omit the constant $\log(2\pi)$ term, since it is a fixed offset on all match costs but would otherwise distort match-vs-miss comparisons. Pairs failing the same $\chi^2_2$ gate as \Cref{sec:clustering} are treated as infeasible ($c_{ij}{=}{+}\infty$).
Three additional terms modify the cost:
\begin{equation}
c_{ij} = c_{ij}^{\mathrm{base}} - \lambda_{\mathrm{assoc}} \cdot \pi^{\mathrm{geo}}_{i,t} - \mu_{\mathrm{sem}} \cdot \cos(\vf_k^{(j)}, \vf_k^{(i)}) + \psi_{\mathrm{turn}}(j, i),
\label{eq:hungarian}
\end{equation}
where $\lambda_{\mathrm{assoc}}$ is the identity boost, $\mu_{\mathrm{sem}}$ is the appearance similarity boost, and $\psi_{\mathrm{turn}}$ is a speed-adaptive turn penalty.

\textbf{Track-level grouping.}
Components are grouped by identity; for each track $t$ and measurement $i$, the grouped cost is $c_{t,i} = \min_{j \in \mathcal{J}_k(t)} c_{ij}$.
We then solve a gated track-level Hungarian assignment over the confirmed-track subset, while tentative/lost hypotheses (including newly birthed tracks) are updated from the remaining measurements in a second pass; the full augmented assignment matrix and second-pass details are given in Supp. Material B. 

Given the resulting assignments, matched components receive a Kalman update:
\begin{align}
K_k^{(j,i)} &= P_{k|k-1}^{(j)} H^\top \bigl(S_k^{(j,i)}\bigr)^{-1}, \label{eq:kalman_gain}\\
\vm_k^{(j)} &= \vm_{k|k-1}^{(j)} + K_k^{(j,i)} (\vz_k^{(i)} - H\vm_{k|k-1}^{(j)}), \label{eq:mean_update}\\
P_k^{(j)} &= (I - K_k^{(j,i)} H)\,P_{k|k-1}^{(j)}\,(I - K_k^{(j,i)} H)^\top + K_k^{(j,i)}\,R_k^{(i)}\,K_k^{(j,i)\top}.
\label{eq:joseph}
\end{align}
These matched components also receive a weight boost $w \leftarrow \min(w+w_{\text{boost}}, 1)$; unmatched components receive a miss penalty $w \leftarrow w \cdot (1 - p_D)$ and are not updated.

\subsubsection{Component Management and Track Lifecycle.}
\label{sec:lifecycle}
After the PHD update, low-weight components are pruned, nearby components with the same identity may be merged, and only the dominant motion-mode component is retained per identity, yielding one representative component per identity for the next frame.

Tracks evolve through \textsc{Tentative}, \textsc{Confirmed}, and \textsc{Lost} states to stabilize persistent IDs and support re-identification after brief occlusion. Exact merge conditions, lifecycle thresholds, deletion rules, and the detailed re-identification score are provided in Supp. Material B. After these transitions, confirmed tracks are extracted and the representative components form the track pool used for the next frame.

\section{Results and Analysis}
\label{sec:experiments}

\subsection{Datasets}

\textbf{WildTrack}~\cite{wildtrack}: 7 synchronized calibrated HD cameras (1920$\times$1080, 2\,fps) covering a 12$\times$36\,m outdoor area with ${\sim}$20 pedestrians per frame; 400 frames total, 360/40 train/test split (90\%/10\%).
\\
\textbf{MultiviewX}~\cite{multiviewx}: 6 synthetic cameras (1920$\times$1080, 2\,fps) covering a 16$\times$25\,m area with ${\sim}$40 pedestrians per frame; same split.
\\
\textbf{RadarScenes}~\cite{radarscenes}: 158 driving sequences from four 77\,GHz radars (72\,868 frames). Each point carries range, azimuth, Doppler velocity, and radar cross-section.


\subsection{Evaluation Metrics}

We report MOT metrics~\cite{bernardin2008clear} (MOTA, MOTP) and identity-aware IDF1~\cite{ristani2016idf1} with a 1.0\,m positive assignment threshold for the camera datasets.
We further report GOSPA~\cite{rahmathullah2017gospa} ($p{=}2$, $c{=}1$, $\alpha{=}2$), which jointly penalizes localization error, missed targets, and false alarms.
For RadarScenes we report LSTQ~\cite{lstq}.

\subsection{Implementation Details}

YOLOv11x~\cite{yolo11_ultralytics} provides monocular object detection; the Lift depth network~\cite{philion2020lift} (EfficientNet-B0, 41 bins) provides an outlier check for fused geometric depth estimates; OSNet-AIN~\cite{zhou2021osnetain} trained with triplet loss via torchreid~\cite{zhou2019torchreid} provides appearance features. All perception modules are trained on the target dataset using the same 90\%/10\% train/test split (e.g., 360/40 frames for WildTrack); Supp. Material E summarizes the exact data preparation, losses, and optimizer settings used for the YOLO, Lift, and OSNet-AIN front-ends. All clustering, tracking, and calibration hyperparameters are detailed in Tables 6 and 7.

\begin{figure}[t]
\centering
\includegraphics[width=0.99\linewidth]{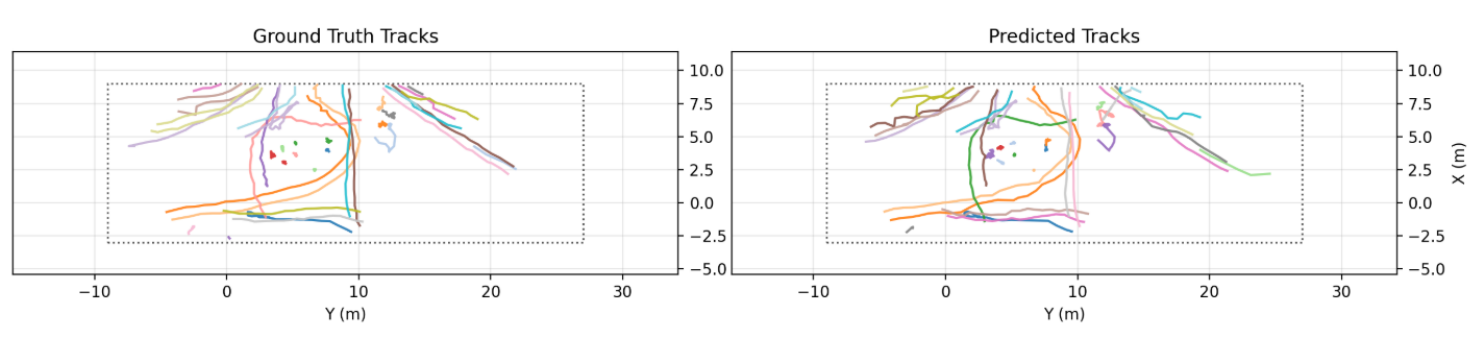}
\caption{Predicted BEV tracks on WildTrack (colored by identity). Ground truth (left) vs. ModTrack (right).}
\label{fig:tracks}
\end{figure}
\subsection{Camera Tracking Results}

\begin{table}[t]
\centering
\caption{Comparison of tracking performance on WildTrack and MultiviewX datasets.}
\label{tab:multiview}
\resizebox{\linewidth}{!}{%
\begin{tabular}{@{}lcccccccc@{}}
\toprule
& \multicolumn{4}{c}{\textbf{WildTrack}} & \multicolumn{4}{c}{\textbf{MultiviewX}} \\
\cmidrule(lr){2-5}\cmidrule(lr){6-9}
Method & IDF1 & MOTA & MOTP & GOSPA$\downarrow$ & IDF1 & MOTA & MOTP & GOSPA$\downarrow$ \\
\midrule
\multicolumn{9}{@{}l}{\emph{End-to-end (fixed config.)}} \\
EarlyBird~\cite{earlybird}     & 92.3 & 89.5 & 86.6 & -- & 82.4 & 88.4 & 86.2 & -- \\
Ali~et~al.~\cite{ali2025directional} & 92.0 & 87.4 & -- & -- & -- & -- & -- & -- \\
MCBLT~\cite{mcblt}              & 95.6 & 92.6 & \textbf{94.3} & -- & --   & --   & --   & -- \\
TrackTacular~\cite{tracktacular}& 95.3 & 91.8 & 85.4 & -- & 85.6 & 92.4 & 80.1 & -- \\
SCFusion~\cite{scfusion}       & 95.9 & 92.4 & 86.3 & -- & 85.0 & 92.5 & 85.4 & -- \\
Alturki~et~al.~\cite{alturki2025attention} & 96.1 & 92.7 & 88.8 & -- & 85.7 & 91.3 & 86.9 & -- \\
MVTrajecter~\cite{mvtrajecter}  & \textbf{96.5} & \textbf{94.3} & 93.0 & -- & \textbf{85.8} & \textbf{92.8} & \textbf{95.0} & -- \\
\midrule
\multicolumn{9}{@{}l}{\emph{Learned (Non-Interpretable)}} \\
DMCT~\cite{dmct}               & 77.8 & 72.8 & 79.1 & -- & --   & --   & --   & -- \\
ReST~\cite{rest}               & 86.7 & 84.9 & 84.1 & -- & --   & --   & --   & -- \\
UMPN~\cite{onegraph}           & 96.3 & 93.9 & 86.9 & -- & --   & --   & --   & -- \\
\midrule
\multicolumn{9}{@{}l}{\emph{Modular (flexible config.)}} \\
KSP-DO~\cite{wildtrack}        & 73.2 & 69.6 & 61.5 & -- & --   & --   & --   & -- \\
MV-GLMB-OC~\cite{ong}         & 74.3 & 69.7 & 73.2 & -- & --   & --   & --   & -- \\
\textbf{ModTrack (Ours)} & \textbf{95.5} & \textbf{91.4} & \textbf{87.2} & \textbf{1.20} & \textbf{86.2} & \textbf{86.3} & \textbf{84.0} & \textbf{1.83} \\
ModTrack (Oracle det.) & 97.1 & 94.7 & 94.0 & 0.68 & 88.2 & 93.6 & 88.8 & 1.25 \\
\bottomrule
\end{tabular}}
\end{table}

\vspace{-1.0em}

\Cref{fig:tracks} shows qualitative tracking results of ModTrack on the test set
of WildTrack.
\Cref{tab:multiview} compares ModTrack against end-to-end and modular baselines on WildTrack and MultiviewX.
Among modular methods, ModTrack surpasses all prior work by $+21.2$ IDF1 and $+21.7$ MOTA on WildTrack.
GOSPA scores of 1.20 and 1.83 reaffirms strong tracking performance.

The best end-to-end method~\cite{mvtrajecter} exceeds ModTrack by 1.0 IDF1 and 2.9 MOTA on WildTrack; in exchange, ModTrack provides sensor-agnostic deployment without retraining and physically traceable uncertainty. 
As for MultiviewX, we investigated the larger gap between ModTrack and other methods (86.3 vs.\ 92.8 MOTA) with a sweep over $\tau_\text{euc} \in \{0.10, 0.25, 0.50, 0.75\}$ which revealed a structural tradeoff: increasing the gate reduces FN clusters (682 to 56) but increases merge errors (225 to 481), with no threshold resolving both. This suggests a structural limitation rather than a tuning issue that emerged with MultiviewX due to its high $2\times$ higher pedestrian density. However, ModTrack leads all methods in IDF1 on MultiviewX, the primary tracking consistency metric, despite the higher crowd density.

\textbf{Oracle detection analysis.} To isolate the quality of the tracker backend from upstream perception noise, we replace the output of the monocular object detector with ground-truth bounding boxes; we refer to this configuration as \textit{oracle detection}. Under oracle detection, ModTrack achieves 97.1/94.7/94.0 (IDF1/MOTA/MOTP) on WildTrack, which would be SOTA. On MultiviewX, oracle performance reaches 88.2/93.6 (IDF1/MOTA), closing the MOTA gap with end-to-end methods entirely (93.6 vs.\ 92.8). These results establish that the residual gap under standard operation originates entirely in the upstream perception modules. Because ModTrack's perception modules are replaceable, improvements in detection or depth estimation directly improve tracking without modifying the tracker.

\subsection{Ablation Studies}

\begin{table}[t]
\centering
\caption{Ablation of matching modes on WildTrack and MultiviewX. Joint mode uses spatial clustering and reserves semantic features for the tracking stage.}
\label{tab:ablation_modes}
\resizebox{.74\linewidth}{!}{%
\begin{tabular}{@{}lcccc|cccc@{}}
\toprule
& \multicolumn{4}{c}{\textbf{WildTrack}} & \multicolumn{4}{c}{\textbf{MultiviewX}} \\
\cmidrule(lr){2-5}\cmidrule(lr){6-9}
Mode & IDF1 & MOTA & MOTP & GOSPA & IDF1 & MOTA & MOTP & GOSPA \\
\midrule
Spatial only           & 86.6 & 89.5 & 82.4 & 1.22 & 84.2 & 84.9 & 83.7 & 1.86 \\
Semantic only          & 49.6 & 41.8 & 79.6 & 2.60 & 65.1 & 56.7 & 79.6 & 3.00 \\
\textbf{Joint (full)}  & \textbf{95.5} & \textbf{91.4} & \textbf{87.2} & \textbf{1.20} &
\textbf{86.2} & \textbf{86.3} & \textbf{84.0} & \textbf{1.83} \\
\bottomrule
\end{tabular}}
\end{table}

\Cref{tab:ablation_modes} compares ModTrack's three matching modes on the two camera datasets: geometric-only association (\textit{Spatial}), appearance-only association (\textit{Semantic}), and the combined \textit{Joint} configuration. Spatial-only achieves strong MOTA (89.5), confirming that calibrated geometric reasoning alone provides robust association. The +8.9 IDF1 improvement from joint mode on WildTrack (+2.0 on MultiviewX) comes entirely from better identity preservation through appearance-informed tracking. Supp. Material Section H ablates over other key design choices. 

\subsection{Deployment Robustness: Sensor Dropout and Addition}

\begin{table}[t]
\centering
\caption{Sensor dropout robustness (WildTrack, joint mode). Mean$\pm$std over all $\binom{7}{k}$ subsets. GOSPA ($p{=}2$, $c{=}1$\,m, $\alpha{=}2$): lower is better.}
\label{tab:dropout}
\resizebox{0.85\linewidth}{!}{%
\begin{tabular}{@{}lcccccc@{}}
\toprule
& \multicolumn{6}{c}{\textbf{Cameras available}} \\
\cmidrule(lr){2-7}
& \makebox[2.8em][l]{\hspace{0.6em}7}
& \makebox[5.8em][l]{\hspace{0.8em}6}
& \makebox[5.8em][l]{\hspace{0.8em}5}
& \makebox[5.8em][l]{\hspace{0.8em}4}
& \makebox[6.4em][l]{\hspace{0.8em}3}
& \makebox[6.4em][l]{\hspace{0.8em}2} \\
\midrule
IDF1  & \makebox[2.8em][l]{95.5} & \makebox[5.8em][l]{91.7$\pm$2.9} & \makebox[5.8em][l]{85.1$\pm$5.3} & \makebox[5.8em][l]{74.5$\pm$9.7} & \makebox[6.4em][l]{57.8$\pm$15.7} & \makebox[6.4em][l]{31.6$\pm$19.5} \\
MOTA  & \makebox[2.8em][l]{91.4} & \makebox[5.8em][l]{87.7$\pm$2.0} & \makebox[5.8em][l]{81.1$\pm$4.7} & \makebox[5.8em][l]{70.2$\pm$9.0} & \makebox[6.4em][l]{51.7$\pm$16.2} & \makebox[6.4em][l]{25.5$\pm$18.3} \\
MOTP  & \makebox[2.8em][l]{87.2} & \makebox[5.8em][l]{85.9$\pm$0.9} & \makebox[5.8em][l]{83.8$\pm$2.9} & \makebox[5.8em][l]{82.4$\pm$3.0} & \makebox[6.4em][l]{80.7$\pm$4.2}  & \makebox[6.4em][l]{73.2$\pm$18.2} \\
GOSPA & \makebox[2.8em][l]{1.2}  & \makebox[5.8em][l]{1.3$\pm$0.1}  & \makebox[5.8em][l]{1.5$\pm$0.2}  & \makebox[5.8em][l]{1.9$\pm$0.3}  & \makebox[6.4em][l]{2.3$\pm$0.4}   & \makebox[6.4em][l]{2.9$\pm$0.4} \\
\bottomrule
\end{tabular}}
\end{table}

\Cref{tab:dropout} quantifies ModTrack's sensor dropout robustness by exhaustively evaluating all $\binom{7}{k}$ camera subsets on WildTrack (120 configurations).

In real-world scenarios where sensor dropout or addition can happen instantaneously, our system adapts seamlessly and effectively, while end-to-end methods in \Cref{tab:multiview} would need additional engineering accommodations or retraining.
With only 5 of 7 cameras, ModTrack achieves 85.1 IDF1 and 81.1 MOTA, surpassing MV-GLMB-OC (74.3/69.7) \emph{with all 7 cameras} by over 10 points. Additionally, MOTP remains above 80 with just 3 cameras, confirming graceful degradation across all metrics as camera coverage reduces.

The increasing standard deviation for IDF1 at lower camera counts ($\pm$2.9 at $k{=}6$ vs.\ $\pm$15.7 at $k{=}3$) quantifies which camera placements are critical, directly informing redundancy planning.

\begin{figure}[t]
\centering
\includegraphics[width=\linewidth]{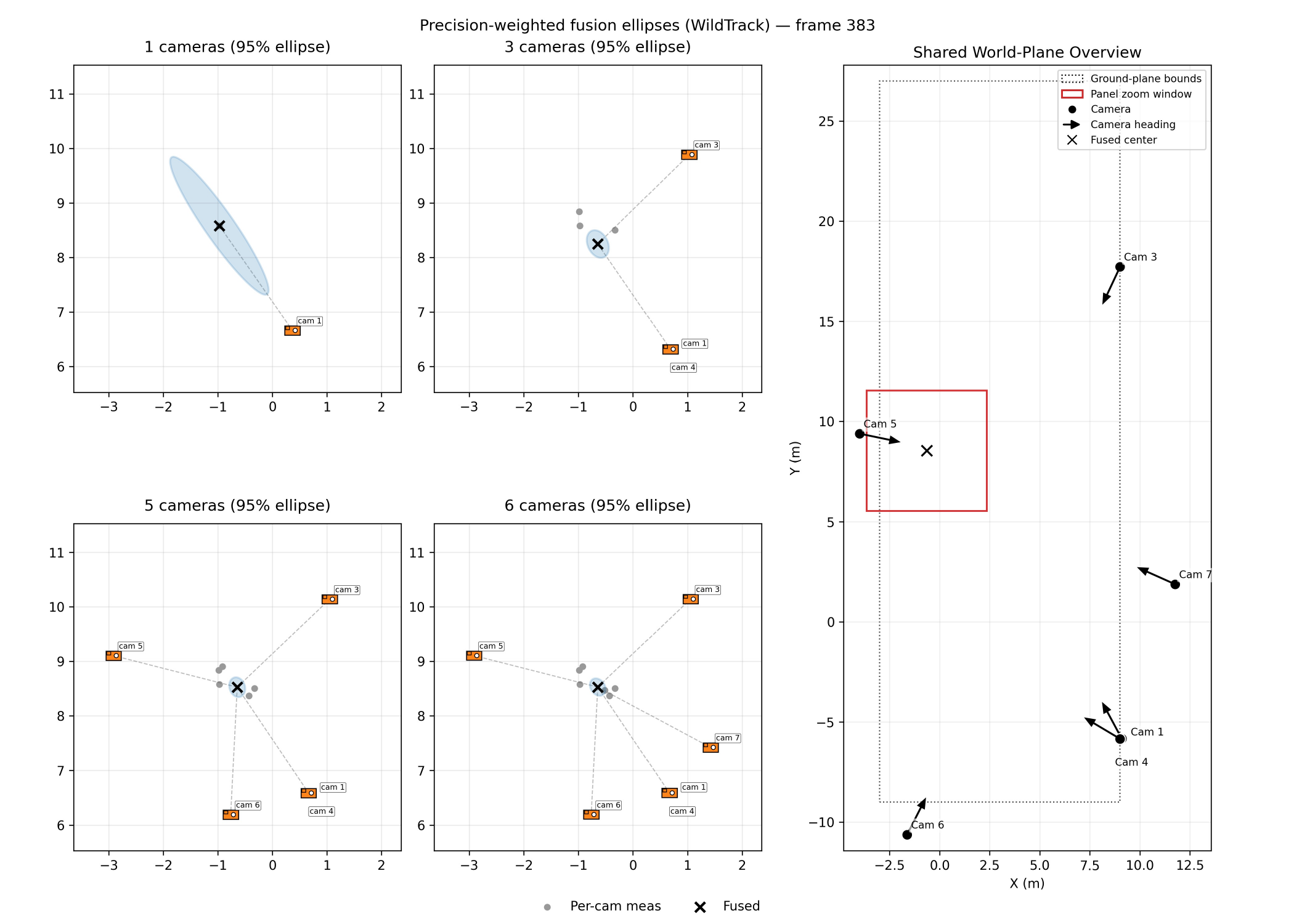}
\caption{Precision-weighted fusion on WildTrack (frame 383). Each panel shows BEV covariance ellipses (95\% confidence) as cameras are incrementally added. The rightmost panel shows the 6-camera shared world-plane.}
\label{fig:covariance}
\end{figure}

\Cref{fig:covariance} visualizes how each additional sensor contributes measurable information through precision-weighted fusion, with shrinking covariance ellipses quantifying the reduction in localization uncertainty that underlies the performance trends in \Cref{tab:dropout}.

\subsection{Radar Tracking Results}
\label{sec:radar-tracking-results}

\begin{table}[!t]
\centering
\caption{Comparison on RadarScenes dataset.}
\label{tab:radar}
\resizebox{.6\linewidth}{!}{%
\begin{tabular}{@{}lcccc@{}}
\toprule
Approach & LSTQ & $S_\text{cls}$ & $S_\text{assoc}$ & GOSPA\\
\midrule
MOT~\cite{radarmot} & 42.4 & 19.4 & 92.7 & -- \\
CA-Net~\cite{canet} & 34.8 & 13.0 & 92.7 & -- \\
Radar Tracker~\cite{radartracker}          & 66.8 & 92.7 & 48.2 & -- \\
ModTrack + Doppler perception  & 42.1 & 70.0 & 26.9 & 2.50 \\
ModTrack + Oracle perception      & 76.9 & 99.1 & 60.9 & 1.09 \\
\bottomrule
\end{tabular}}
\end{table}
The $(\vz, R)$ interface extends naturally to any sensor producing a calibrated BEV position and covariance.
We validate this by applying the identical tracker core to automotive radar on RadarScenes~\cite{radarscenes}, replacing only the perception front-end (\Cref{sec:radar_frontend}).

\Cref{tab:radar} shows ModTrack achieves 42.1 LSTQ with a zero-training Doppler perception front-end. Providing the ModTrack with Oracle ground-truth labels raises LSTQ to 76.9 and surpasses the best performing method reported in the literature (Radar Tracker) by 10.1 points.
The entire gap is in perception: a stronger front-end translates directly into better tracking without any changes to the tracker.

\subsection{Runtime Analysis}
\label{sec:runtime}
When tested on WildTrack, the tracking backend accounts for less than 20\% of per-frame latency (103\,ms vs.\ 444\,ms for neural perception on an NVIDIA RTX 3090 GPU); a full breakdown is provided in Supp. Material C.

\subsection{Sensitivity Analysis}
To assess the robustness of ModTrack to hyperparameter choices, we perform a one-at-a-time sensitivity analysis on the WildTrack benchmark.
\label{sec:sensitivity}

\begin{figure}[t]
    \centering
    \begin{subfigure}[b]{0.48\textwidth}
        \centering
        \includegraphics[width=\textwidth]{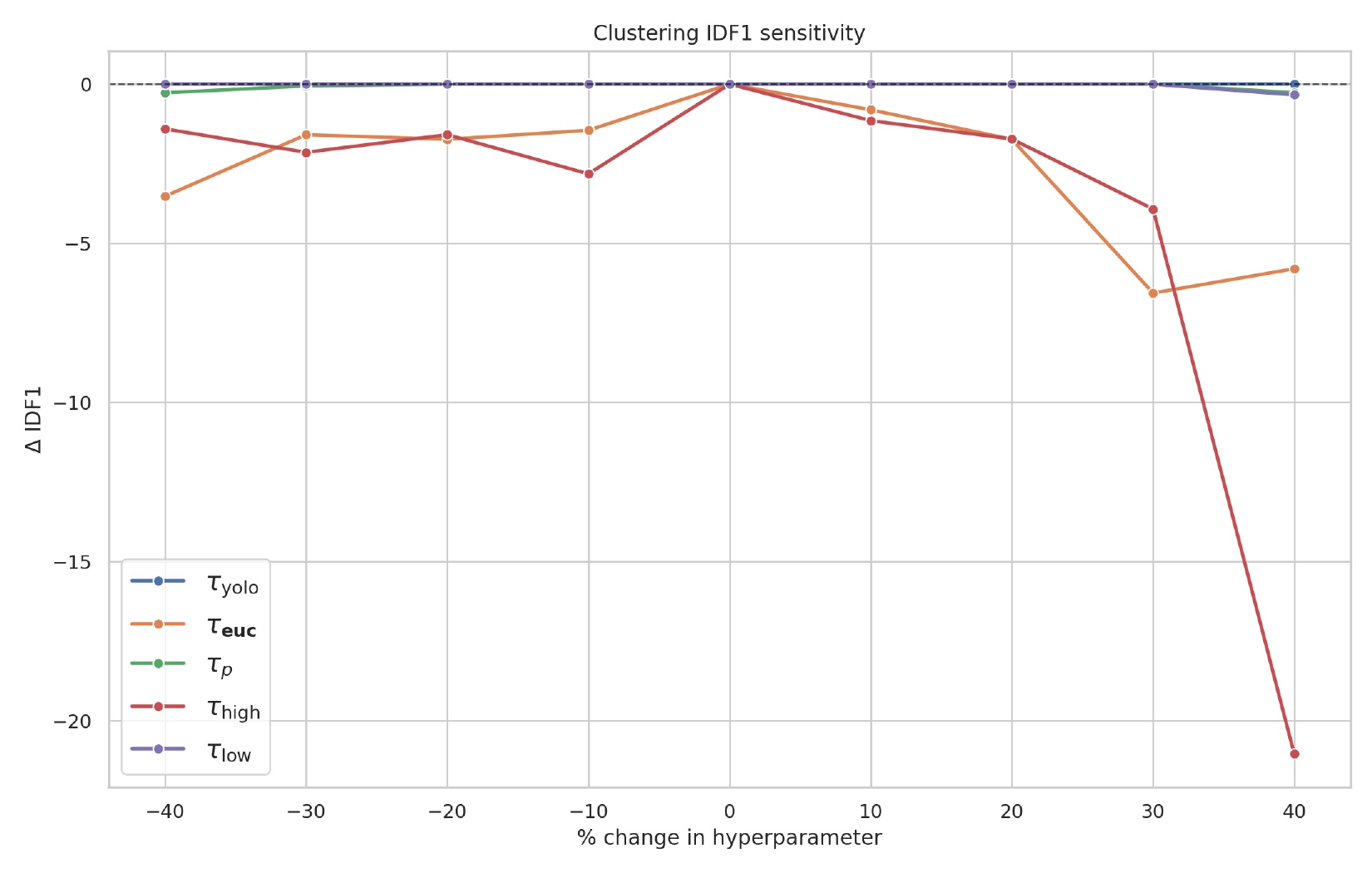}
        \caption{Clustering hyperparameters.}
        \label{fig:clustering_idf1}
    \end{subfigure}
    \hfill
    \begin{subfigure}[b]{0.48\textwidth}
        \centering
        \includegraphics[width=\textwidth]{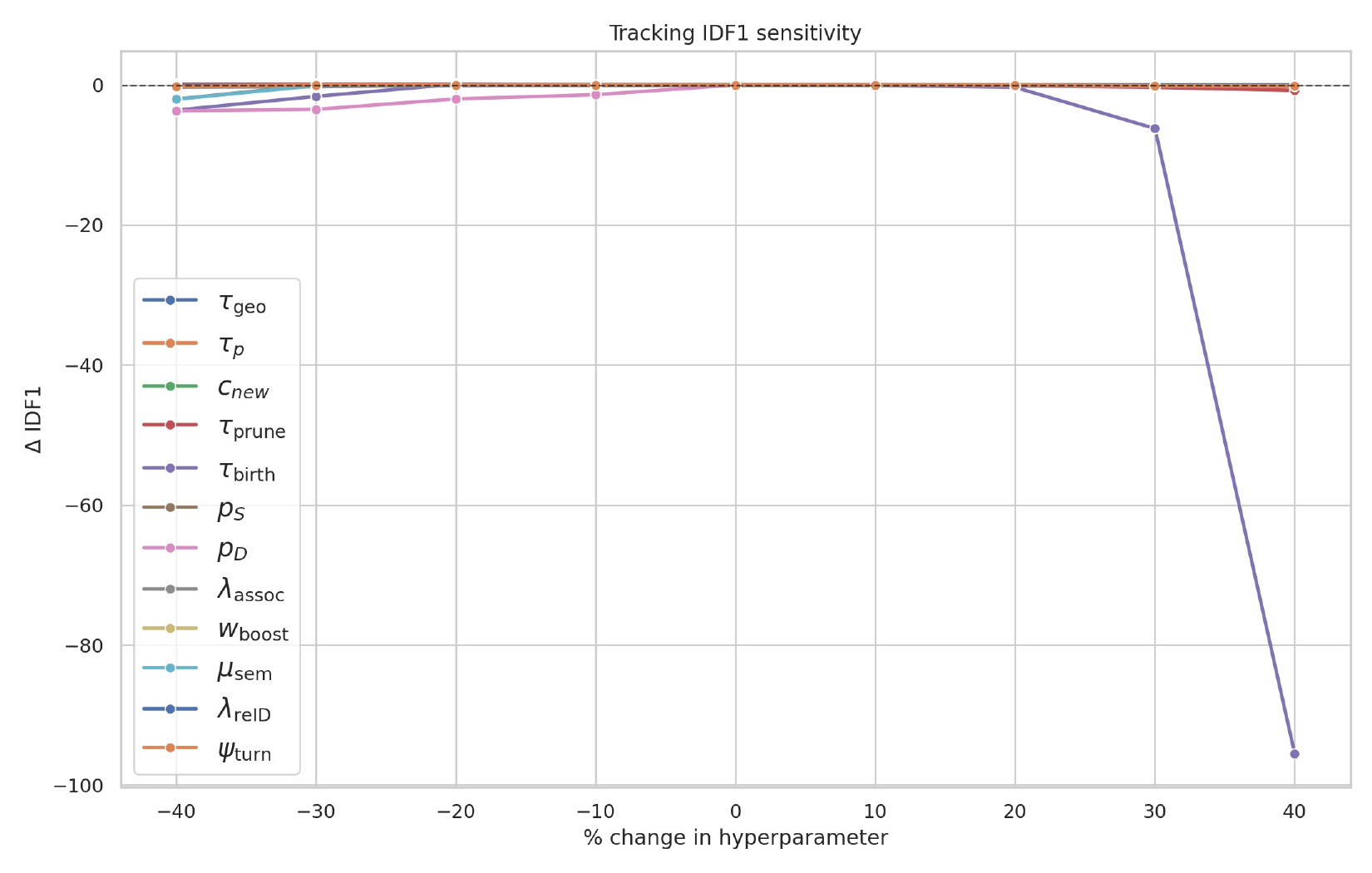}
        \caption{Continuous tracking hyperparameters.}
        \label{fig:tracking_idf1}
    \end{subfigure}
    \caption{IDF1 sensitivity analysis (WildTrack, joint mode). Each parameter is swept individually; all others held at defaults. Extended results are in Supp. Material D.}
    \label{fig:sensitivity_main}
\end{figure}

Only three parameters, $\tau_{\mathrm{euc}}$ (Euclidean gating), $\tau_{\mathrm{high}}$ (ByteTrack high confidence threshold), and $\tau_{\mathrm{birth}}$ (minimum cluster confidence for track birth), meaningfully affect IDF1 (\Cref{fig:sensitivity_main}); all three directly control the FP/FN tradeoff, and all remaining parameters vary by at most 2--3 IDF1 points.

\section{Conclusion}
\label{sec:conclusion}

ModTrack demonstrates that restricting neural networks to perception while performing downstream reasoning with closed-form analytical methods can match end-to-end performance while providing deployment flexibility, auditable uncertainty, and sensor agnosticism. Across WildTrack, MultiviewX, and RadarScenes, the same identity-informed GM-PHD tracking core operates unchanged, while the closed-loop identity mechanism and sensor-agnostic $(\vz, R)$ covariance chain enable robust operation under sensor dropout. Oracle detection experiments confirm that the tracker backend is not the limiting factor: with improved perception, ModTrack's tracking performance scales directly toward SOTA. More broadly, the explicit propagation of calibrated uncertainty throughout the pipeline enables future sensor placement and selection strategies that maximize information gain in detection and tracking systems.

\section*{Acknowledgements}
This work was supported by the Office of Naval Research (ONR) Award No. N00014-24-1-2662
\bibliographystyle{splncs04}
\bibliography{modtrack}
\newpage
\appendix

\section*{Supplementary Material}

\section{Depth Estimation Details}
\label{app:depth}

\textbf{Planar footpoint projection.}
The bottom-center pixel $(u_f, v_f)$ of each bounding box is back-projected using camera intrinsics $K$ to obtain the normalized camera-frame ray
\[
\mathbf{r}_c = K^{-1}[u_f, v_f, 1]^\top.
\]
Assuming a known ground plane $\mathbf{n}^\top \mathbf{X}_w = d_{\mathrm{plane}}$ and camera-to-world extrinsics $(R_c, \mathbf{t}_c)$, the signed ray--plane intersection parameter is
\[
\lambda_{\mathrm{fp}} =
\frac{d_{\mathrm{plane}} - \mathbf{n}^\top \mathbf{t}_c}
         {\mathbf{n}^\top (R_c\,\mathbf{r}_c)}.
\]
We define the world/BEV frame so the ground plane is horizontal, i.e., $\mathbf{n} = [0,0,1]^\top$ and $\mathbf{n}^\top \mathbf{X}_w = z_w = d_{\mathrm{plane}}$. Thus, the orthogonal projection in Section 3.2 changes only $z_w$, so \Cref{eq:jacobian} needs no additional in-plane projection factor.

We define the metric depth estimate as
\begin{equation}
d_{\mathrm{fp}} = |\lambda_{\mathrm{fp}}|, \quad
\sigma^2_{\mathrm{fp}} =
\max\!\left((\alpha_{fp}\,d_{\mathrm{fp}})^2,\,
\sigma^2_{\min\mathrm{depth}}\right),
\label{eq:footpoint}
\end{equation}
where $\alpha_{fp}=0.035$ controls relative geometric uncertainty.

\textbf{Bounding-box depth prior.}
A complementary estimate follows from the observed bounding-box height $h_{\mathrm{px}}$ under the pinhole model with vertical focal length $f_y$ and reference object height $H_{\mathrm{ref}}$:
\begin{equation}
d_{\mathrm{bbox}} =
\frac{f_y\,H_{\mathrm{ref}}}{h_{\mathrm{px}}}, \quad
\sigma^2_{\mathrm{bbox}} =
\max\!\left((\alpha_{bbox}\,d_{\mathrm{bbox}})^2,\,
\sigma^2_{\min\mathrm{depth}}\right),
\label{eq:bbox_prior}
\end{equation}
where $\alpha_{bbox}=0.05$ controls the relative bounding-box uncertainty.

\textbf{Fused depth estimate.}
The two geometric cues are combined via precision weighting:
\begin{equation}
\hat{d} =
\frac{\pi_{\mathrm{fp}}\,d_{\mathrm{fp}} +
      \pi_{\mathrm{bbox}}\,d_{\mathrm{bbox}}}
     {\pi_{\mathrm{fp}} + \pi_{\mathrm{bbox}}},
\quad
\pi_{\mathrm{fp}} =
\frac{\eta_{\mathrm{fp}}}{\sigma^2_{\mathrm{fp}}},
\quad
\pi_{\mathrm{bbox}} =
\frac{1}{\sigma^2_{\mathrm{bbox}}},
\label{eq:depth_fusion}
\end{equation}
where $\eta_{\mathrm{fp}}=3$ reflects the lower systematic bias of the ground-plane cue. Depth uncertainty is parameterized as
\[
\sigma_d^2 = \max\!\left((\alpha_{fp}\,|\hat{d}|)^2,\;\sigma^2_{\min\mathrm{depth}}\right).
\]
Finally, a Lift-style depth network~\cite{philion2020lift} is used as an outlier check: if its predicted depth differs from $|\hat{d}|$ by more than $3\,\mathrm{m}$ while indicating high confidence, the variance is inflated by $\gamma_{\mathrm{inflate}}=1.75$.

\subsection{BEV Projection and Covariance Propagation}
\label{app:bev_projection}
Section 3.2 gives the operational summary; here we provide the full projection and covariance derivation used for camera measurements.

Given a detection at pixel $(u, v)$ in camera $c$, we back-project
$\mathbf{r}_c = K^{-1}[u, v, 1]^\top$ and form the world point
\begin{equation}
\mathbf{X}_w = R_c\,(\hat{d}\,\mathbf{r}_c) + \mathbf{t}_c.
\end{equation}
To enforce a ground-plane measurement, we orthogonally project onto
$\mathbf{n}^\top \mathbf{X} = d_{\mathrm{plane}}$:
\begin{equation}
\mathbf{X}_{\mathrm{proj}} = \mathbf{X}_w + \left(d_{\mathrm{plane}} - \mathbf{n}^\top \mathbf{X}_w\right)\mathbf{n},
\end{equation}
and define the BEV position $\vz_{\mathrm{BEV}} = \mathbf{X}_{\mathrm{proj},\,1:2}$.

Holding $(u,v)$ fixed, the full derivative through the plane projection is $(I - \mathbf{n}\mathbf{n}^\top)\,R_c\,\mathbf{r}_c$; since we define $\mathbf{n} = [0,0,1]^\top$ (horizontal ground plane, see \Cref{app:depth}), the projector $(I - \mathbf{n}\mathbf{n}^\top) = \mathrm{diag}(1,1,0)$ affects only the vertical component, so extracting rows 1:2 gives the Jacobian
\begin{equation}
J = \frac{\partial \vz_{\mathrm{BEV}}}{\partial \hat{d}}
= R_{c,\,1:2,:}\,K^{-1}\begin{bmatrix}u\\v\\1\end{bmatrix}
\in \R^{2\times 1}.
\label{eq:jacobian}
\end{equation}
Propagating depth uncertainty yields the rank-1 BEV covariance term
\[
R_{\mathrm{indep}} = \sigma_d^2\,J\,J^\top.
\]

We then add the isotropic pose/calibration term
\[
\Rpose = \sigma_{\mathrm{pose}}^2 I_2.
\]
To preserve a minimum eigenvalue $\sigma^2_{\min}$ for the final BEV covariance, we floor only the independent depth term before adding $\Rpose$. With $\lambda_{\min}(\Rpose)$ the smallest eigenvalue of $\Rpose$, we set
\begin{equation}
\sigma^2_{\min,\mathrm{indep}} = \max\!\left(0,\,\sigma^2_{\min} - \lambda_{\min}(\Rpose)\right),
\end{equation}
and clamp the eigenvalues of $R_{\mathrm{indep}}$ to be at least $\sigma^2_{\min,\mathrm{indep}}$. The final measurement covariance is
\[
\Rbev = R_{\mathrm{indep}} + \Rpose.
\]

The $\Rbev = R_{\mathrm{indep}} + \Rpose$ decomposition is physically motivated: $R_{\mathrm{indep}}$ averages down with sensor count while $\Rpose$ does not shrink with additional sensors. The fixed-covariance ablation in Section~\ref{sec:choice_ablation} supports this: replacing the structured form with fixed isotropic $R\!=\!0.16I$ drops IDF1 (95.5 to 95.0).

\section{Full Identity-Informed GM-PHD Details}
\label{app:phd_details}

\subsection{Association and Update Details}
\textbf{Track-level grouping.}
Components are grouped by identity; for each track $t$ and measurement $i$, the grouped cost is $c_{t,i} = \min_{j \in \mathcal{J}_k(t)} c_{ij}$. Collecting these entries yields the track-level cost matrix $C_{\mathrm{track}} = [c_{t,i}]$.

Dedicated miss columns with cost $=-\log(1 - p_D^{(c)})$ allow tracks to remain unmatched:
\begin{equation}
C_{\mathrm{full}} = \bigl[\, C_{\mathrm{track}} \;\big|\; \mathrm{diag}(-\log(1 - p_D^{(c)}))\,\bigr].
\label{eq:augmented_cost}
\end{equation}
Hungarian assignment on $C_{\mathrm{full}}$ yields globally optimal 1:1 track-to-measurement matching.

We apply \Cref{eq:augmented_cost} only to \textsc{Confirmed} tracks to avoid weight dilution in dense scenes. Remaining \textsc{Tentative}/\textsc{Lost} tracks are then processed in a second pass using only measurements not claimed by the confirmed-track Hungarian: we accumulate soft association evidence over $\chi^2_2$-gated pairs (using $d_{ij}^2$ from Equation 16 (main paper)) and apply the Kalman update in Equation 20 (main paper) only for a best match that satisfies a tighter gate ($\chi^2_{2,0.90}=4.61$). For \textsc{Lost} tracks, we additionally boost re-identification using appearance similarity as in Section 3.5.

\subsection{Component Management and Lifecycle Details}
After the PHD update, component management is applied before any tracks are exposed externally. Low-weight components are pruned first, then components with identical identities and small Mahalanobis distance ($< \tau_{\mathrm{merge}}$) are merged via moment matching, and we finally retain only the highest-weight motion-mode component per identity, yielding one representative component (track state) per persistent ID for lifecycle handling and next-frame association.

Tracks follow a three-state lifecycle: \textsc{Tentative} $\to$ \textsc{Confirmed} (after $N_{\mathrm{init}}$ hits or deleted if misses exceed $\tau_{\mathrm{tent}}$) $\to$ \textsc{Lost} (after $\tau_{\mathrm{confirmed}}$ consecutive misses), with deletion after age $K_{\max}$ without re-association.
Hit counters decay gradually ($h \leftarrow \max(0, h{-}1)$) rather than resetting on a miss, preserving accumulated evidence.
Re-identification of lost tracks uses a combined spatial-semantic score: $\mathrm{score}(i,j) = -d^2_{\mathrm{Mahal}} + \lambda_{\mathrm{reID}} \cdot \cos(\vf_{\mathrm{lost}}, \vf_{\mathrm{meas}})$.

\section{Runtime Analysis}
\label{app:runtime}

\begin{table}[H]
\centering
\caption{Per-frame inference time breakdown on WildTrack (7 cameras, RTX 3090). The analytical core accounts for less than 20\% of total latency; perception dominates runtime.}
\label{tab:runtime}
\small
\setlength{\tabcolsep}{8pt}
\begin{tabular}{@{}lrrr@{}}
\toprule
\textbf{Stage} & \textbf{Time (ms)} & \textbf{\% Total} & \textbf{FPS} \\
\midrule
Neural perception (Det.\ + Depth + Feat.) & 443.9 & 81.0 & 2.25 \\
Analytical core (Clust.\ + Fusion + PHD) & 103.2 & 18.8 & 9.69 \\
Other overhead & 0.7 & 0.1 & -- \\
\midrule
\textbf{ModTrack (Total)} & \textbf{547.8 $\pm$ 85.0} & \textbf{100} & \textbf{1.83} \\
\bottomrule
\end{tabular}
\end{table}

\Cref{tab:runtime} reports per-frame timing on WildTrack (RTX 3090), averaged after skipping five warm-up frames. MCBLT~\cite{mcblt} reports a 1.5 FPS speed for their tracking system, whereas our method runs at 1.83 FPS, as shown in \Cref{tab:runtime}.

\section{Extended Sensitivity Analysis}
\label{app:sensitivity}

This Supplementary Material provides additional sensitivity results beyond the clustering and continuous tracking sweeps shown in Figure 4 (main paper).
Each hyperparameter is swept while all others are held at their default values.

\begin{figure}[h]
    \centering
    \includegraphics[width=0.65\textwidth]{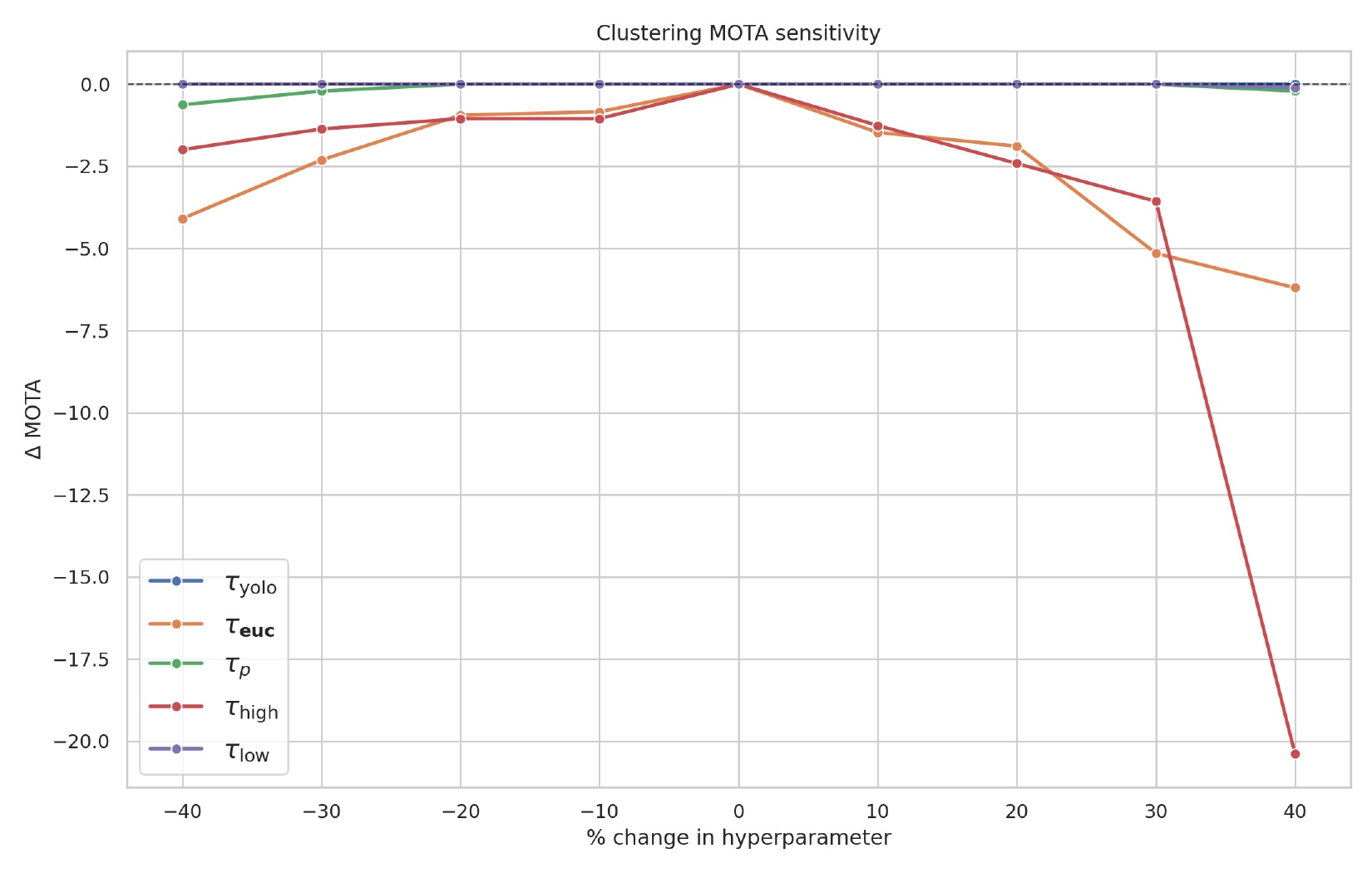}
    \caption{MOTA sensitivity to continuous clustering hyperparameters on WildTrack.}
    \label{fig:clustering_mota}
\end{figure}

\begin{figure}[h]
    \centering
    \includegraphics[width=0.65\textwidth]{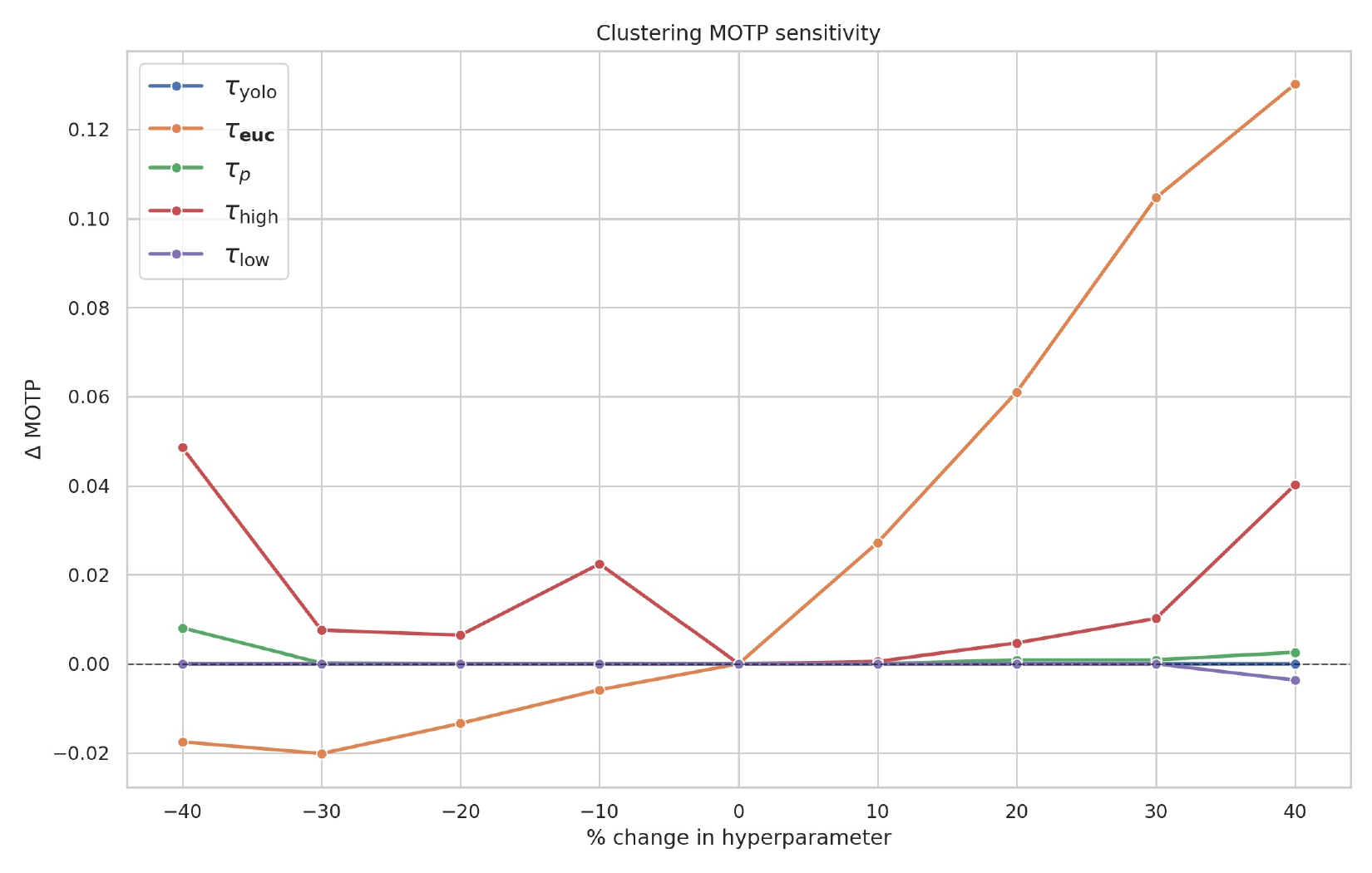}
    \caption{MOTP sensitivity to continuous clustering hyperparameters on WildTrack.}
    \label{fig:clustering_motp}
\end{figure}

\begin{figure}[h]
    \centering
    \includegraphics[width=0.65\textwidth]{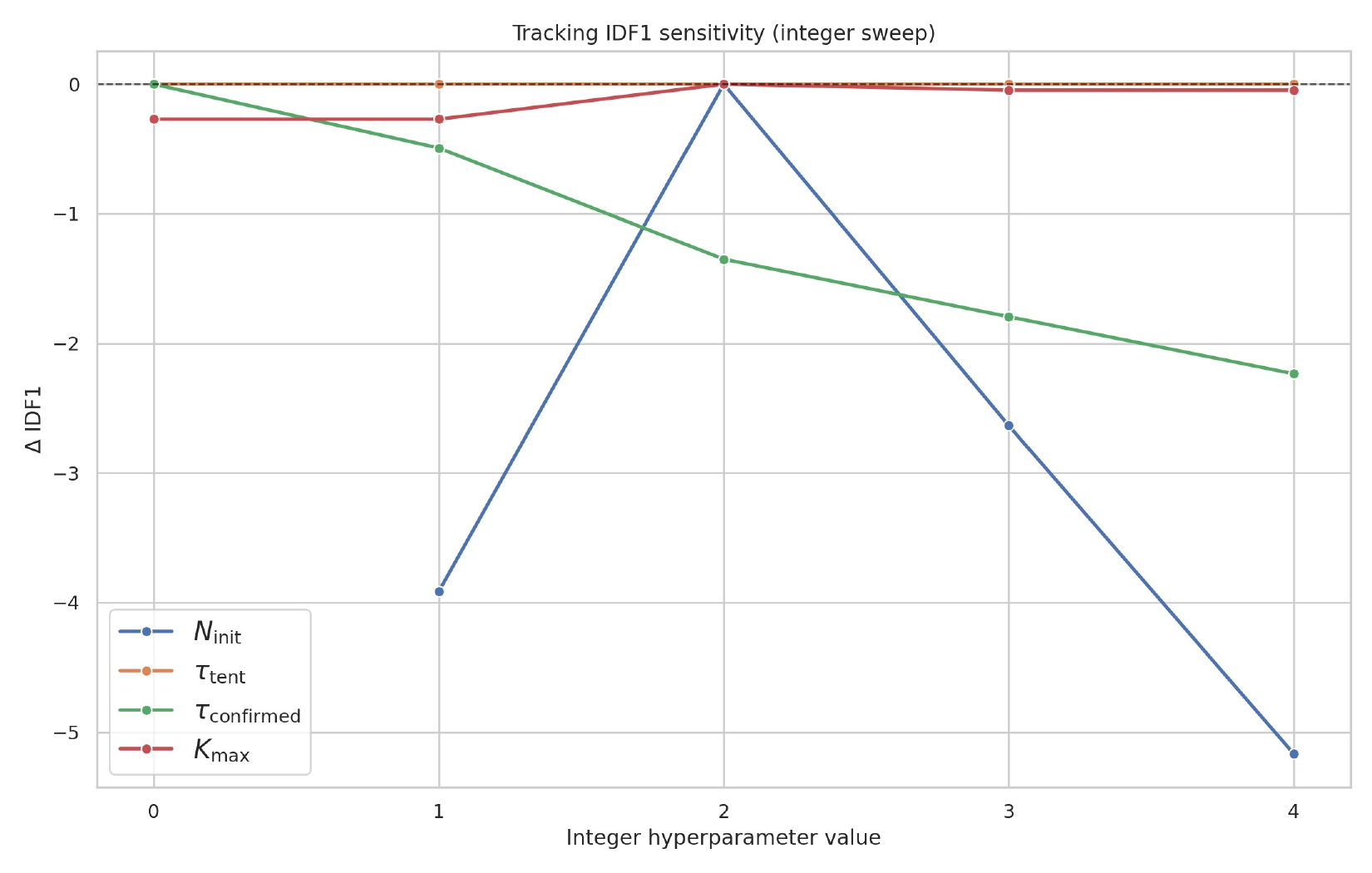}
    \caption{IDF1 sensitivity to integer tracking hyperparameters ($N_{\mathrm{init}}$, $\tau_{\mathrm{tent}}$, $\tau_{\mathrm{confirmed}}$, $K_{\max}$) on WildTrack.}
    \label{fig:tracking_idf1_int}
\end{figure}

\begin{figure}[H]
    \centering
    \includegraphics[width=0.65\textwidth]{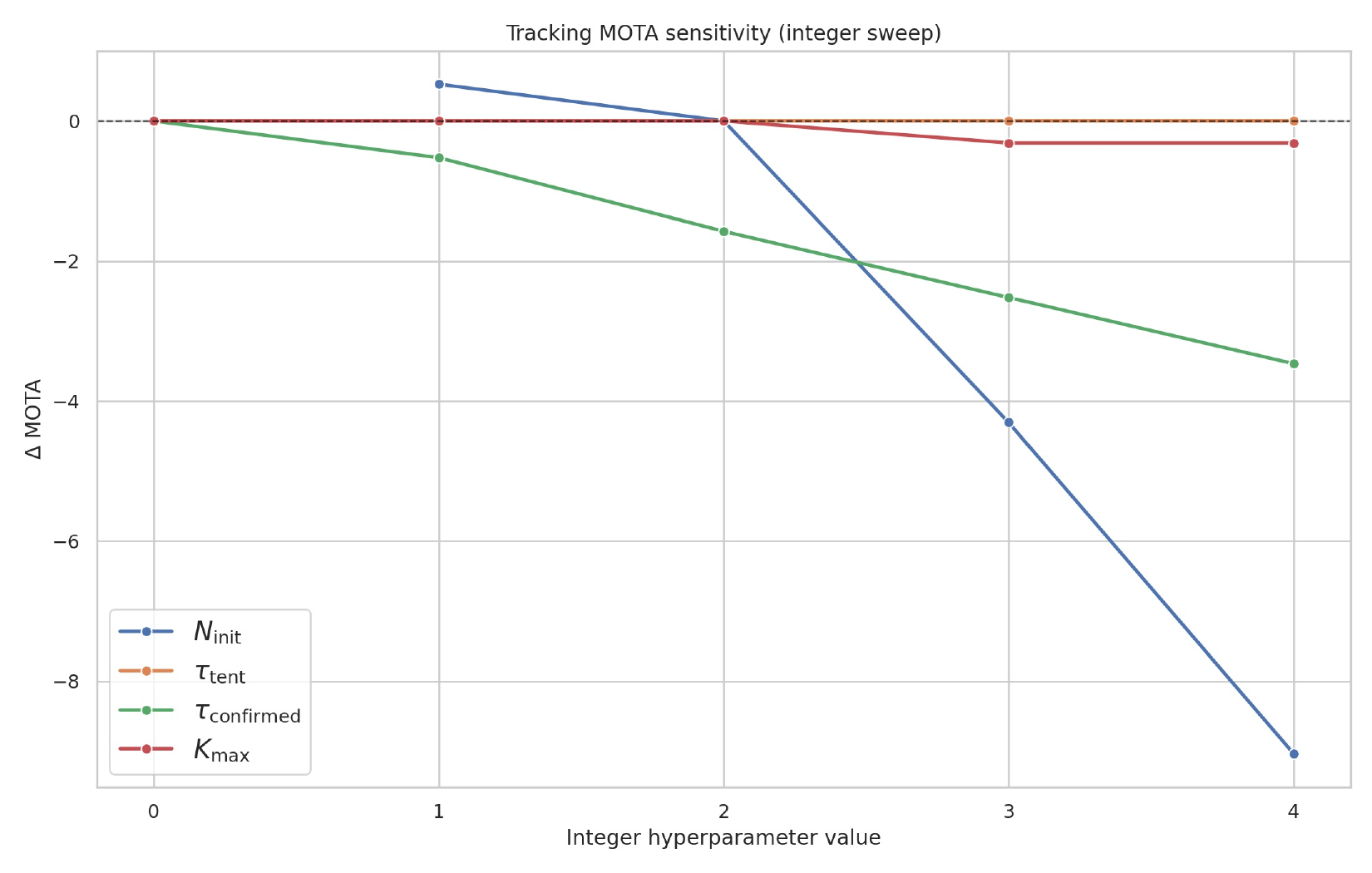}
    \caption{MOTA sensitivity to integer tracking hyperparameters ($N_{\mathrm{init}}$, $\tau_{\mathrm{tent}}$, $\tau_{\mathrm{confirmed}}$, $K_{\max}$) on WildTrack.}
    \label{fig:tracking_mota_int}
\end{figure}

\begin{figure}[h]
    \centering
    \includegraphics[width=0.65\textwidth]{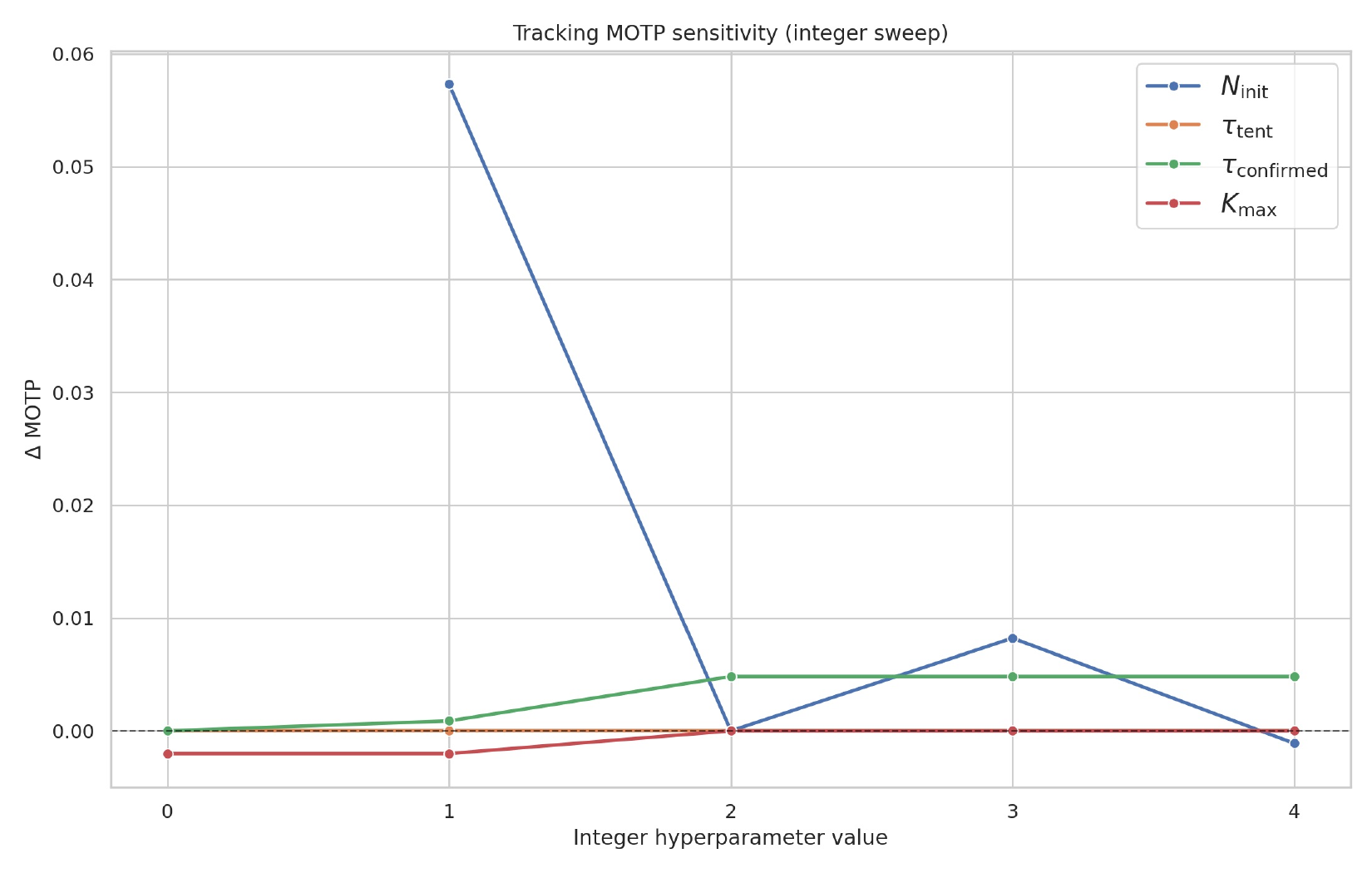}
    \caption{MOTP sensitivity to integer tracking hyperparameters ($N_{\mathrm{init}}$, $\tau_{\mathrm{tent}}$, $\tau_{\mathrm{confirmed}}$, $K_{\max}$) on WildTrack.}
    \label{fig:tracking_motp_int}
\end{figure}

\begin{figure}[h]
    \centering
    \includegraphics[width=0.65\textwidth]{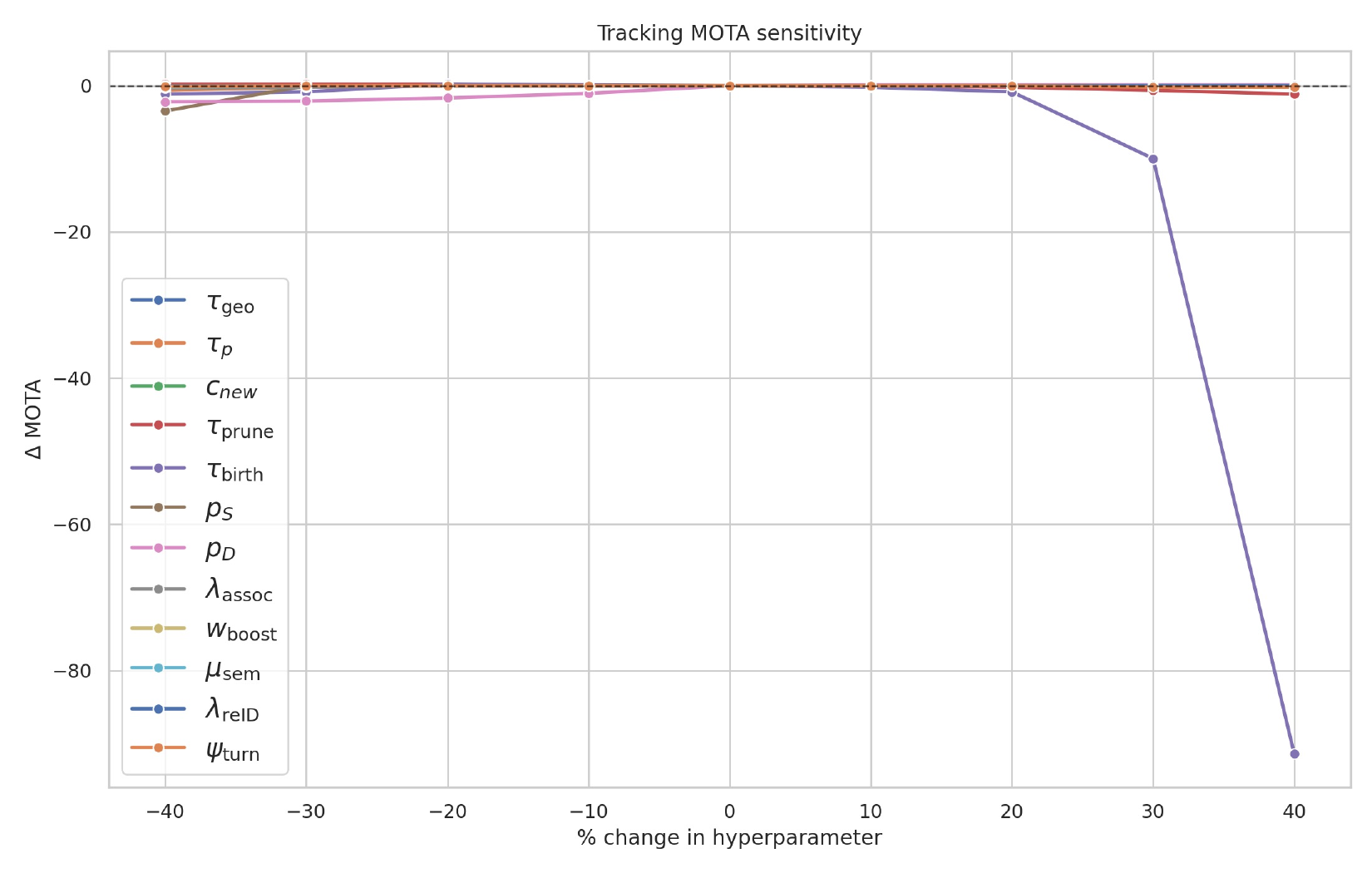}
    \caption{MOTA sensitivity to continuous tracking hyperparameters on WildTrack.}
    \label{fig:tracking_mota}
\end{figure}

\begin{figure}[h]
    \centering
    \includegraphics[width=0.65\textwidth]{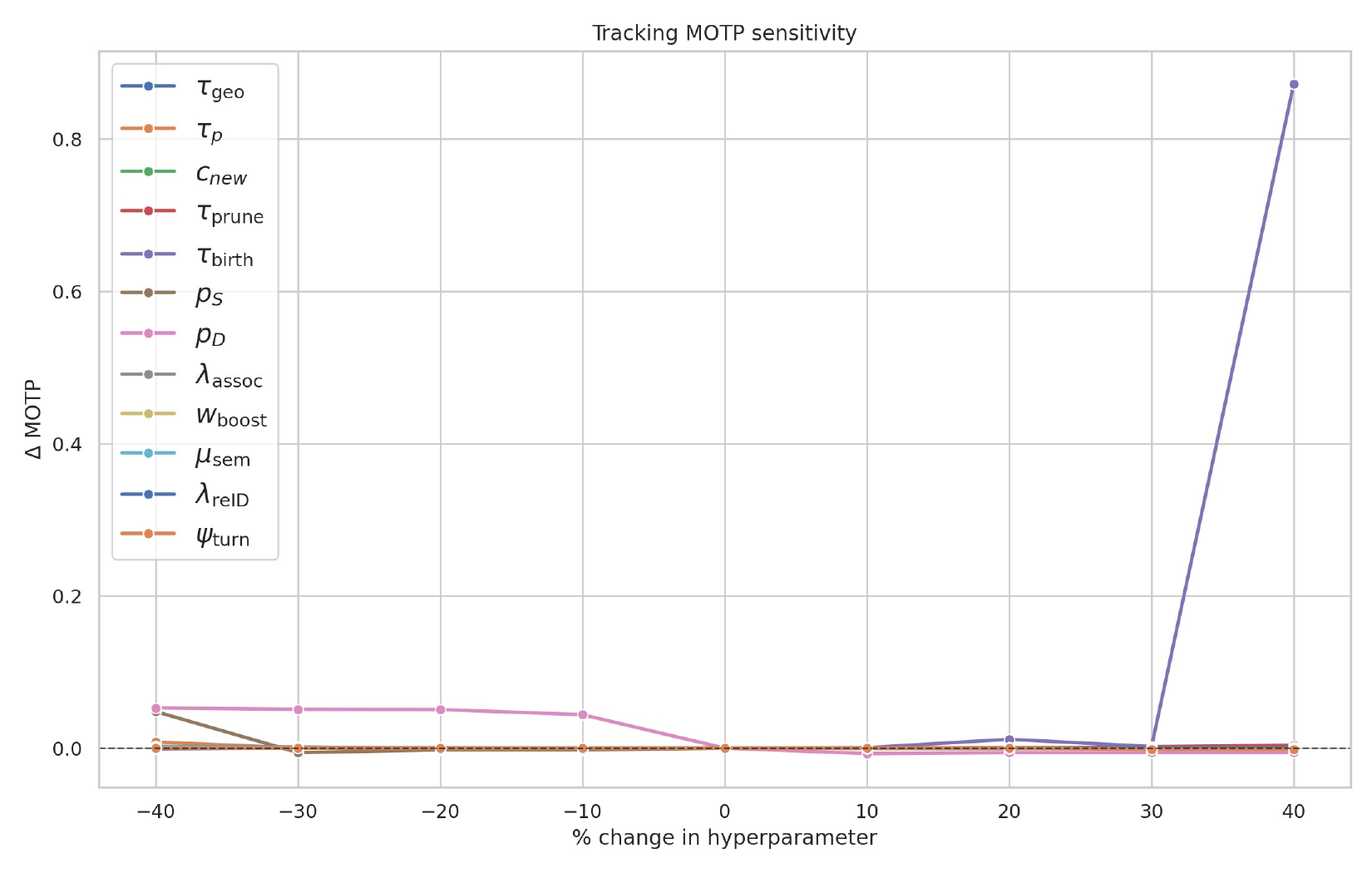}
    \caption{MOTP sensitivity to continuous tracking hyperparameters on WildTrack.}
    \label{fig:tracking_motp}
\end{figure}

\begin{figure}[h]
    \centering
    \includegraphics[width=0.65\textwidth]{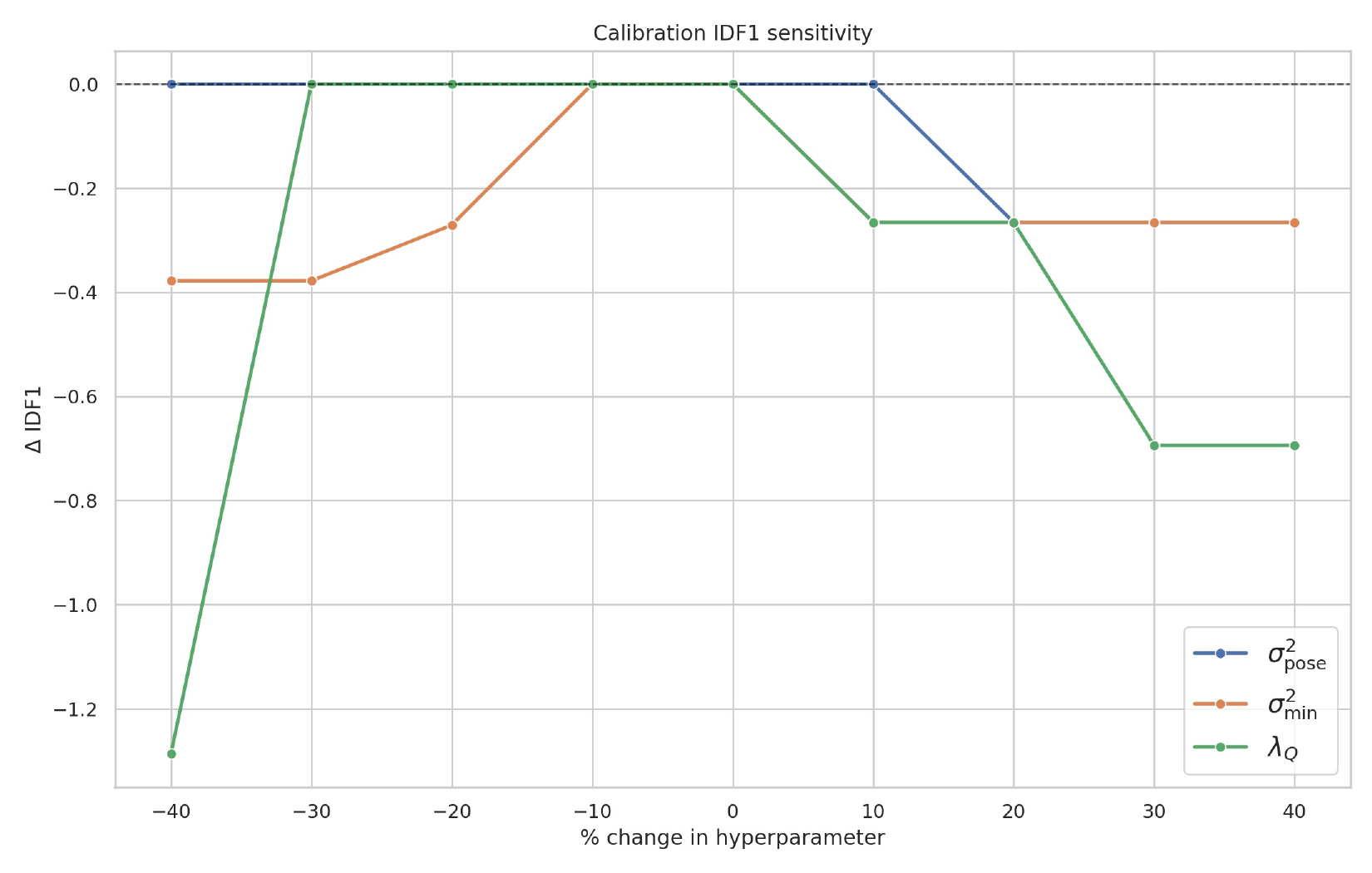}
    \caption{IDF1 sensitivity to calibration hyperparameters ($\sigma^2_{\mathrm{pose}}$, $\sigma^2_{\mathrm{min}}$, $\lambda_Q$) on WildTrack.}
    \label{fig:calib_idf1}
\end{figure}

\begin{figure}[h]
    \centering
    \includegraphics[width=0.65\textwidth]{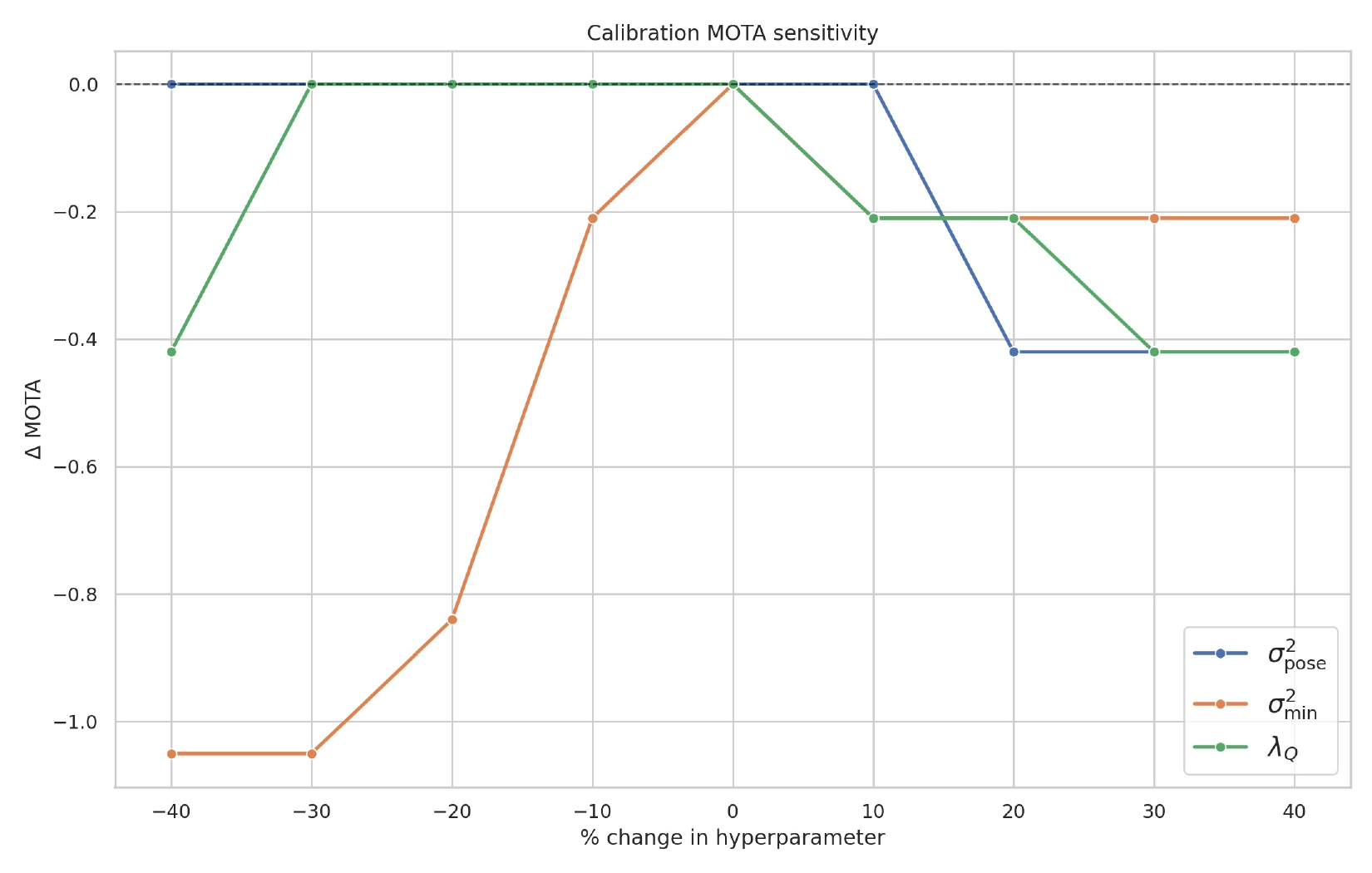}
    \caption{MOTA sensitivity to calibration hyperparameters ($\sigma^2_{\mathrm{pose}}$, $\sigma^2_{\mathrm{min}}$, $\lambda_Q$) on WildTrack.}
    \label{fig:calib_mota}
\end{figure}

\begin{figure}[h]
    \centering
    \includegraphics[width=0.65\textwidth]{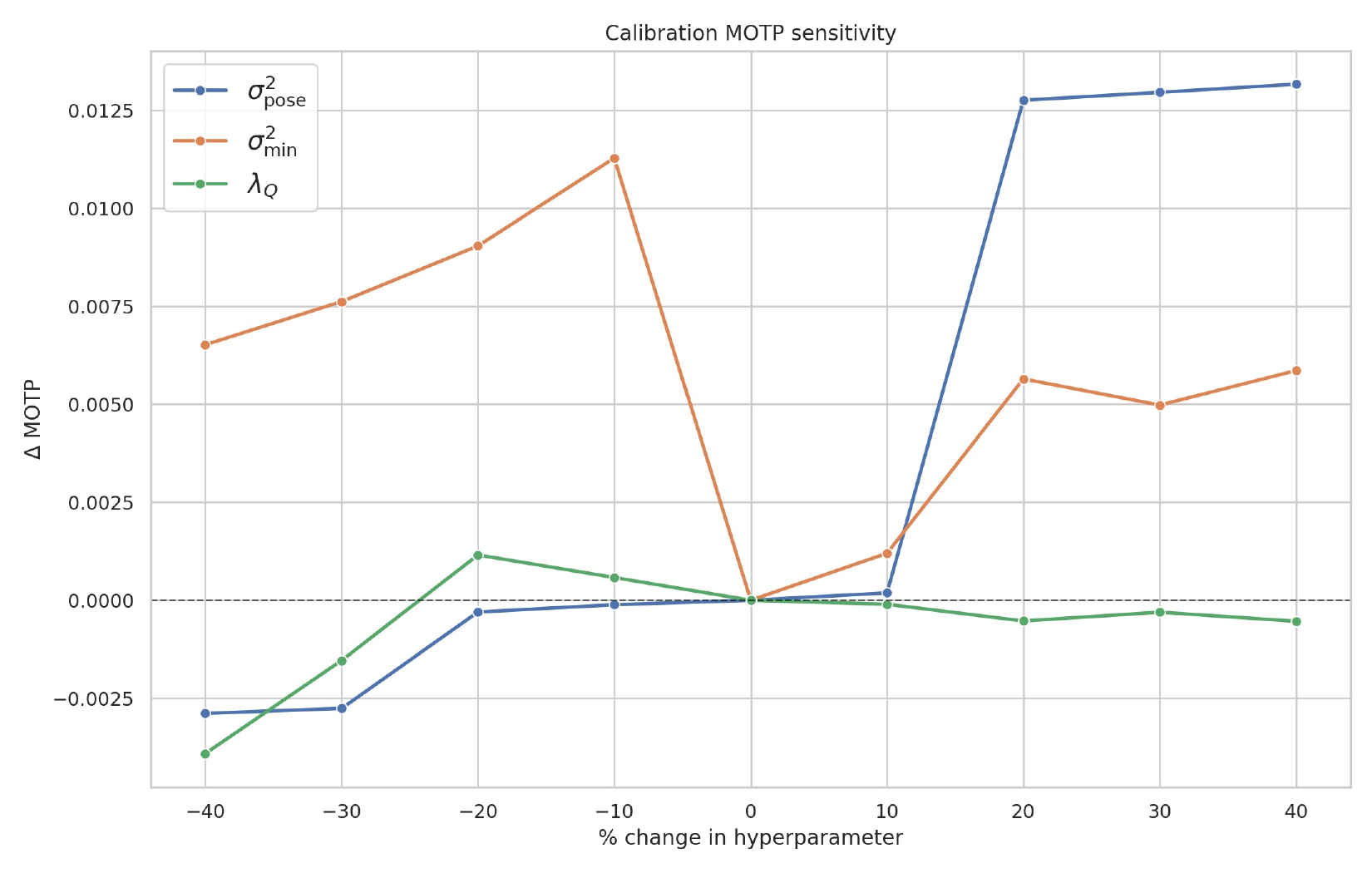}
    \caption{MOTP sensitivity to calibration hyperparameters ($\sigma^2_{\mathrm{pose}}$, $\sigma^2_{\mathrm{min}}$, $\lambda_Q$) on WildTrack.}
    \label{fig:calib_motp}
\end{figure}

\textbf{Clustering parameters} (Figure 4a (main paper), and \Cref{fig:clustering_mota,fig:clustering_motp}).

\textbf{Tracking parameters} (\Cref{fig:tracking_idf1_int,fig:tracking_mota_int,fig:tracking_motp_int,fig:tracking_mota,fig:tracking_motp} and Figure 4b (main paper)).
$N_{\mathrm{init}}$ (hits required for confirmation) and $\tau_{\mathrm{confirmed}}$ (misses before entering lost state) show moderate sensitivity but only to the effect of ${\sim}$2--3 percentage points of IDF1.
$K_{\max}$ (maximum lost age before deletion) is largely insensitive and $\tau_{\mathrm{tent}}$ has no effect within the tested range.

\textbf{Calibration parameters} (\Cref{fig:calib_idf1,fig:calib_mota,fig:calib_motp}). All calibration parameters show minimal sensitivity in the range of ~1.2 points of IDF1, ~1.0 points of MOTA, and ~0.0025 points of MOTP.

\section{Camera Perception Training Details}
\label{app:perception_training}

All three learned front-ends are trained independently on a single NVIDIA RTX 3090 GPU using the same ordered 90/10 frame split in the main experiments for WildTrack and MultiviewX. This keeps the perception stack aligned with the tracker evaluation split while avoiding frame-level leakage across time.

\subsection{YOLOv11x Detector Fine-Tuning}
The YOLO model is fine-tuned with the standard YOLO training pipeline maintained by Ultralytics~\cite{yolo11_ultralytics}, using \texttt{imgsz}$=640$ so the network sees a consistent square training resolution. We initialize from a pretrained YOLOv11x checkpoint, train for 50 epochs with batch size 8. After training, the best validation checkpoint is retained as the dataset-specific detector weights.

\subsection{Lift Depth Network Training}
Ground truth for the depth model is obtained by assigning the annotated 3D person location to every pixel within its corresponding ground-truth 2D box, yielding sparse but geometrically exact supervision. The CamEncode network is initialized from the released LSS checkpoint, uses 41 linear depth bins over $[1,36]$\,m, a context dimension of 64, and $360{\times}640$ RGB inputs; images are scaled to [0,1] and normalized with ImageNet mean/std to match the EfficientNet-B0 backbone initialization. 
Training runs for 50 epochs with batch size 8 using AdamW, learning rate $10^{-4}$, weight decay $10^{-4}$, and gradient clipping at 1.0. 
During training the model predicts depth logits over the full image, for each ground truth box a Gaussian centered in the box is used to sample $n=16$ points, the expected depth is computed from the predicted sampled depth logits, and an $L_1$ loss is applied against the box's ground truth metric depth. 
The checkpoint with the lowest validation $L_1$ error is retained as the final Lift model.

\subsection{OSNet-AIN ReID Training}
The appearance encoder is trained on cropped person patches organized by identity and camera. Only identities observed in at least two cameras are retained so the learned embedding is explicitly cross-view. Training uses a PK sampler: with the default batch size of 64, each mini-batch contains 16 identities and four images per identity. Images are resized to $256{\times}128$ and augmented with padding, random crop, horizontal flip, color jitter, normalization, and random erasing. OSNet-AIN is trained with batch-hard triplet loss (margin 0.3). We train for 150 epochs and 350 PK batches per epoch using AdamW (learning rate $3{\times}10^{-4}$, weight decay $5{\times}10^{-4}$), cosine annealing to $10^{-6}$, and gradient clipping at 5.0. Model selection is based on the validation \emph{pair-gap} statistic, i.e., the mean separation between same-identity and different-identity cosine-similarity rates across the chosen thresholds, and the checkpoint with the highest pair-gap is used at inference time.

\section{Hyperparameter Summary}
\label{app:hyperparams}

\begin{table}[t!]
\centering
\caption{Hyperparameters: perception, projection, and clustering stages. Camera-specific parameters (top block) do not apply to RadarScenes.}
\label{tab:hyperparams}
\small
\resizebox{\linewidth}{!}{%
\begin{tabular}{@{}llcccp{4.8cm}@{}}
\toprule
\textbf{Stage} & \textbf{Symbol} & \textbf{WildTrack} & \textbf{MultiviewX} & \textbf{RadarScenes} & \textbf{Physical Interpretation} \\
\midrule
\multicolumn{6}{@{}l}{\emph{Camera Perception}} \\
& $\tau_{\text{yolo}}$ & 0.1 & 0.63 & -- & Min.\ detector confidence to retain a detection \\
& $\alpha_{fp}$ & 0.035 & 0.035 & -- & Footpoint depth uncertainty scaling factor \\
& $\alpha_{bbox}$ & 0.05 & 0.05 & -- & Bbox depth uncertainty scaling factor \\
& $\eta_{\mathrm{fp}}$ & 3 & 3 & -- & Footpoint precision boost (trust factor) \\
& $\sigma^2_{\min\mathrm{depth}}$ & 0.0001 & 0.0001 & -- & Min.\ depth variance \\
& $\gamma_{\mathrm{inflate}}$ & 1.75 & 1.75 & -- & Variance inflation when Lift disagrees \\
\midrule
\multicolumn{6}{@{}l}{\emph{Radar Perception}} \\
& $\sigma_r$ & -- & -- & 0.10\,m & Range measurement std.\ dev. \\
& $\sigma_\theta$ & -- & -- & 1.8$^\circ$ & Azimuth measurement std.\ dev. \\
& $\epsilon_{\mathrm{DBSCAN}}$ & -- & -- & 2.5\,m & DBSCAN neighborhood radius \\
& $n_{\min}$ & -- & -- & 2 & DBSCAN min.\ samples per cluster \\
& $v_{\min}$ & -- & -- & 0.5\,m/s & Min.\ Doppler speed (moving-object filter) \\
\midrule
\multicolumn{6}{@{}l}{\emph{BEV Projection}} \\
& $\sigma_{\mathrm{pose}}$ & 0.17\,m & 0.22\,m & 0.32\,m & Calibration/ego-pose uncertainty ($\sqrt{\Rpose}$) \\
& $\sigma_{\min}^2$ & 0.16\,m$^2$ & 0.21\,m$^2$ & 0.21\,m$^2$ & Min.\ eigenvalue floor on BEV covariance \\
\midrule
\multicolumn{6}{@{}l}{\emph{Clustering}} \\
& $\chi^2$ gate & 9.21 & 9.21 & 13.82 & $\chi^2_2$ quantile (99\% camera, 99.9\% radar) \\
& $\tau_{\mathrm{euc}}$ & 0.5\,m & 0.5\,m & 0.5\,m & Hard Euclidean cutoff to prevent chain effects \\
& $\tau_{\text{high}}$ & 0.5 & 0.63 & -- & ByteTrack high-confidence threshold \\
& $\tau_{\text{low}}$ & 0.2 & -- & -- & ByteTrack low-confidence threshold \\
\bottomrule
\end{tabular}}
\end{table}

\begin{table}[t!]
\centering
\caption{Hyperparameters: tracking, lifecycle, and motion model stages. RadarScenes uses the same GM-PHD tracking loop with dataset-specific tuning.}
\label{tab:hyperparams2}
\small
\resizebox{\linewidth}{!}{%
\begin{tabular}{@{}llcccp{4.8cm}@{}}
\toprule
\textbf{Stage} & \textbf{Symbol} & \textbf{WildTrack} & \textbf{MultiviewX} & \textbf{RadarScenes} & \textbf{Physical Interpretation} \\
\midrule
\multicolumn{6}{@{}l}{\emph{Tracking (GM-PHD)}} \\
& $p_S$ & 0.99 & 0.99 & 0.99 & Target survival probability \\
& $p_D$ & 0.90 & 0.95 & 0.90 & Detection probability \\
& $\lambda_{\mathrm{assoc}}$ & 2.5 & 2.5 & 2.5 & Identity score boost in assignment cost \\
& $\mu_{\mathrm{sem}}$ & 2.0 & 2.0 & 0.0 & Appearance similarity boost (spatial-only for radar) \\
& $\lambda_{\mathrm{ReID}}$ & 3.0 & 3.0 & 0.0 & Appearance boost for LOST-track re-identification \\
& $w_{\text{boost}}$ & 0.15 & 0.15 & 0.15 & Weight boost for matched components \\
& $\sigma_{\mathrm{spatial}}$ & 1.0 & 1.0 & 1.0 & Spatial bandwidth for identity kernel \\
& $\tau_{\mathrm{geo}}$ & 5.0 & 5.0 & 5.0 & Softmax temperature for identity score \\
& $\tau_{\mathrm{new}}$ & 12.0 & 12.0 & 12.0 & Cost of spawning new identity in front-end Hungarian \\
& $\sigma_v^{(c)}$ & 1.0\,m/s & 1.0\,m/s & 8.0\,m/s & Birth velocity prior std.\ dev. \\
& $\tau_{\mathrm{birth}}$ & 0.65 & 0.65 & 0.40 & Min.\ cluster confidence for track birth \\
& $Q_{\mathrm{scale}}$ & 0.9 & 0.9 & 15.0 & Process noise scaling factor \\
\midrule
\multicolumn{6}{@{}l}{\emph{Track Lifecycle}} \\
& $N_{\mathrm{init}}$ & 2 & 2 & 2 & Hits to confirm a tentative track \\
& $K_{\max}$ & 2 & 2 & 5 & Max.\ lost age before deletion \\
& $J_{\max}$ & 100 & 100 & 100 & Max.\ mixture components retained after capping \\
& $\tau_{\mathrm{confirmed}}$ & 0 & 1 & 3 & Consecutive misses before entering lost state \\
& $\tau_{\mathrm{tent}}$ & 1 & 0 & 2 & Consecutive tentative misses before deletion \\
& $\tau_{\mathrm{prune}}$ & 0.05 & 0.05 & 0.05 & Min.\ component weight during pruning \\
& $\tau_{\mathrm{merge}}$ & 2.5 & 2.5 & 2.5 & Mahalanobis threshold for merging components \\
\midrule
\multicolumn{6}{@{}l}{\emph{HMM Motion Modes}} \\
& $\pi_{\mathrm{stay}}^{(1)}$ & 0.75 & 0.75 & 0.70 & Stationary mode self-transition prob. \\
& $\pi_{\mathrm{stay}}^{(2)}$ & 0.94 & 0.94 & 0.95 & Constant-velocity mode self-transition prob. \\
& $\pi_{\mathrm{stay}}^{(3)}$ & 0.10 & 0.10 & 0.15 & Maneuvering mode self-transition prob. \\
& $\Delta t$ & 0.5\,s & 0.5\,s & 0.2\,s & Time step between frames \\
\midrule
\multicolumn{6}{@{}l}{\emph{Turn Penalty ($\psi_{\mathrm{turn}}$)}} \\
& \multicolumn{5}{l}{Speed-adaptive penalty: $\psi_{\mathrm{turn}}(j,i) = \lambda_{\mathrm{turn}} \cdot \|\dot{\vx}_k^{(j)}\| \cdot (1 - \cos\Delta\theta_{ji})$,} \\
& \multicolumn{5}{l}{where $\Delta\theta_{ji}$ is the angle between predicted velocity and measurement direction,} \\
& \multicolumn{5}{l}{and $\lambda_{\mathrm{turn}} = 1.5$. Penalizes sharp turns proportionally to speed.} \\
\midrule
\multicolumn{6}{@{}l}{\emph{RadarScenes Post-Processing}} \\
& $d_{\mathrm{merge}}$ & -- & -- & 6.0\,m & Tube merging max distance \\
& $g_{\mathrm{merge}}$ & -- & -- & 5 & Tube merging max frame gap \\
\bottomrule
\end{tabular}}
\end{table}

\Cref{tab:hyperparams,tab:hyperparams2} provide a comprehensive reference for all hyperparameters used in ModTrack across all three datasets. Camera-specific perception parameters (top block of \Cref{tab:hyperparams}) do not apply to RadarScenes, which instead uses the radar-specific parameters listed below them. All clustering, fusion, and tracking parameters (bottom blocks) apply to every dataset, with only numerical values differing.

\section{Uncertainty Calibration Analysis}
\label{sec:appendix_calibration}

We evaluate the statistical calibration of ModTrack's reported BEV state
covariances using the Normalized Estimation Error Squared (NEES).
For each frame, predicted tracks are matched to ground truth via Hungarian
assignment under a 1.0\,m gating threshold. Calibration is therefore evaluated
\emph{conditioned on successful associations}.

For each matched pair, we compute
\[
\text{NEES} = e^\top P^{-1} e,
\]
where $e = \mathbf{z}_{gt} - \mathbf{z}_{pred}$ and $P$ is the predicted
$2\times2$ position covariance. Under correct calibration,
$\mathbb{E}[\text{NEES}] = 2$ and NEES follows a $\chi^2(2)$ distribution.

\paragraph{Primary test (mean NEES).}
Let $N$ be the number of matched samples.
Under the null hypothesis of correct calibration, the mean NEES lies within
\[
\left[
\frac{\chi^2_{2N,\,0.025}}{N},
\frac{\chi^2_{2N,\,0.975}}{N}
\right].
\]
We classify uncertainty as:
\textbf{CALIBRATED} (mean within CI),
\textbf{OVERCONFIDENT} (mean above CI),
or \textbf{CONSERVATIVE} (mean below CI).

\begin{table}[H]
\centering
\small
\setlength{\tabcolsep}{6pt}
\caption{Calibration statistics (matched track/GT pairs only).}
\label{tab:calibration_results}
\begin{tabular}{lccccc}
\toprule
Dataset & $N$ & Mean NEES & 95\% CI  & 1$\sigma$ Cov. & 2$\sigma$ Cov. \\
\midrule
WildTrack & 879  & 0.46 & 1.87-2.13 & 88.4\% & 98.6\% \\
MultiviewX & 1317 & 0.38 & 1.89-2.11 & 91.9\% & 99.6\% \\
\bottomrule
\end{tabular}
\end{table}

\paragraph{Coverage diagnostics.}
In addition to the mean NEES test, we report empirical 1$\sigma$ and
2$\sigma$ coverage. For a 2D Gaussian state, a prediction is inside the
$k\sigma$ ellipse if $\text{NEES} \le k^2$.
Under correct calibration, the expected coverage is
$\mathbb{P}(\chi^2_2 \le k^2)$, which equals 39.3\% for $k=1$
and 86.5\% for $k=2$.

\paragraph{Interpretation.}
On both datasets, mean NEES lies significantly below the theoretical
expectation of 2.0, and empirical coverage exceeds nominal
$\chi^2$ expectations (39.3\% for 1$\sigma$, 86.5\% for 2$\sigma$).
This indicates that ModTrack's propagated covariances are
\emph{conservative}, i.e., slightly over-dispersed relative to empirical
localization error. Importantly, no evidence of overconfidence is observed.
The consistent behavior across datasets suggests that the analytical
covariance propagation and precision-weighted fusion yield stable,
auditable uncertainty estimates.

\section{Design Choice Ablation}
\label{sec:choice_ablation}

We ran component ablations on the Wildtrack dataset to isolate the following core design choices in the ModTrack pipeline : (i) clustering, (ii) covariance propagation, (iii) precision weighted fusion.

\begin{itemize}
    \item \textbf{Clustering (RANSAC).} We replace $\chi^2$-based graph clustering with a RANSAC clustering. Clusters are formed by grouping detections within a fixed Euclidean radius (0.5\,m) in BEV space, enforcing at most one detection per camera per cluster. Unlike standard randomized RANSAC, our implementation is deterministic and exhaustive: for each clustering call, all remaining detections are considered as candidate seeds in a fixed (deterministically shuffled) order, and the best consensus set is selected before removing its members and repeating. This process continues until all detections are assigned, ensuring that results are not sensitive to iteration counts or stochastic sampling. This removes covariance-aware statistical gating and relies purely on geometric agreement.

    \item \textbf{Uncertainty propagation (fixed covariance).} We keep all detection positions unchanged but replace each detection covariance with a constant isotropic matrix ($R = 0.16I$), removing both depth-dependent uncertainty and shared calibration terms. This isolates the effect of uncertainty modeling while preserving measurement scale.

    \item \textbf{Fusion (simple mean).} We retain the same clusters and per-camera selection, but replace precision-weighted fusion with uniform averaging ($\hat{z} = \frac{1}{M}\sum z_i$ and $R = \frac{1}{M}\sum R_i$), treating all detections as equally certain.
\end{itemize}

\begin{table}[H]
\centering
\small
\begin{tabular}{lccc}
\toprule
\textbf{Method} & \textbf{Changed Component} & \textbf{IDF1} & \textbf{MOTA} \\
\midrule
ModTrack (baseline) & None & 95.5 & 91.4 \\
+ RANSAC clustering & Clustering & 92.8 & 90.8 \\
+ Fixed covariance ($R=0.16I$) & Detection covariance & 95.0 & 90.7 \\
+ Simple mean fusion & Fusion & 93.0 & 89.5 \\
\bottomrule
\end{tabular}
\caption{Ablation study isolating key components of ModTrack on WildTrack (joint setting).}
\end{table}

\noindent Each ablation modifies a single component while keeping all other stages (detections, projection, tracking, and evaluation protocol) identical:

These results show that each component contributes meaningfully. In particular, replacing $\chi^2$ clustering and precision-weighted fusion leads to clear degradation, highlighting the importance of uncertainty-aware association and fusion. The smaller drop under fixed covariance indicates that while the system remains robust, explicit uncertainty modeling provides gains.

\section{Long Term Tracking}

To evaluate ModTrack's performance for longer term tracking we ran inference across all 400 WildTrack frames using the perception module that was finetuned on the 90/10 train/test split. Data leakage was intentional to isolate the tracker's performance and ensure degraded perception due to insufficient training data was not a factor. In this setting, ModTrack achieves MOTA 79.3, IDF1 83.0, and GOSPA 1.63. Despite the extended sequence length, performance remains stable with no evidence of compounding error, i.e., progressive degradation from accumulated tracking mistakes such as increasing ID switches over time. This supports the claim that the tracker itself does not inherently deteriorate over longer horizons.

\section{Empirical Validation of Formal RFS/PHD Departure}

Our \textbf{identity-informed GM-PHD} maintains GM-PHD prediction, Kalman update, and covariance propagation, but departs from RFS theory: component weights no longer equal expected target mass, and identities are assigned deterministically (Hungarian) rather than marginalized as in GLMB/LMB. 

We demonstrate that this loss of formal guarantees does not degrade the practical behavior expected of a well-behaved multi-target estimator. Under full sensor coverage, ModTrack substantially improves identity consistency over a vanilla GM-PHD baseline while preserving set estimation accuracy (positions + \# objects) (Table~\ref{tab:rebuttal_results}). 

\begin{table}[H]
\centering
\small
\setlength{\tabcolsep}{4pt}
\begin{tabular}{lcccc}
\toprule
\textbf{Dataset} & \textbf{Method} & \textbf{IDF1} $\uparrow$ & \textbf{IDS} $\downarrow$ & \textbf{GOSPA} $\downarrow$ \\
\midrule
WildTrack & Vanilla & 79.3 & 30 & 1.31 \\
          & ModTrack & \textbf{95.5} & \textbf{2} & \textbf{1.20} \\
\midrule
MultiviewX & Vanilla & 46.2 & 145 & 1.95 \\
           & ModTrack & \textbf{85.2} & \textbf{14} & \textbf{1.83} \\
\bottomrule
\end{tabular}
\caption{Comparison of ModTrack against standard GM-PHD filter on WildTrack.}
\label{tab:rebuttal_results}
\end{table}

Here, vanilla GM-PHD refers to the standard unlabeled GM-PHD filter, which fully preserves the original RFS/PHD formulation.

On WildTrack, ModTrack improves over vanilla GM-PHD from 79.3 to 95.5 IDF1 and improves GOSPA from 1.31 to 1.20. On MultiviewX, IDF1 improves from 46.2 to 85.2 and improves GOSPA from 1.95 to 1.83. This indicates that identity augmentation improves temporal consistency without degrading spatial accuracy or cardinality estimation.

\section{Additional Radar Analysis}
\label{app:radar_extra}
We tested an additional MLP perception module that achieves (LSTQ=36.0, GOSPA=1.42), highlighting that a simple perception module struggles with radar data.

\section{Code Availability}
\label{app:code}

The full codebase is on GitHub. \href{https://github.com/JackRobs25/ModTrack}{https://github.com/JackRobs25/ModTrack}

\clearpage
\end{document}